\documentclass[sigconf,balance=false]{acmart}
\usepackage{popets}

\setcopyright{popets}
\copyrightyear{YYYY}

\acmYear{YYYY}
\acmVolume{YYYY}
\acmNumber{X}
\acmDOI{XXXXXXX.XXXXXXX}
\acmISBN{}
\acmConference{Proceedings on Privacy Enhancing Technologies}
\usepackage{xspace}

\usepackage{tabularx}

\usepackage{amsthm}

\usepackage{amsmath}
\usepackage{dsfont}
\usepackage{wrapfig}
\usepackage[table]{colortbl}
\usepackage{makecell}

\usepackage{caption}
\usepackage{multirow}

\usepackage{graphicx}
\usepackage{textcomp}
\usepackage{colortbl}
\usepackage{standalone}
\usepackage{booktabs}
\usepackage[shortlabels]{enumitem}

\definecolor{LightCyan}{rgb}{0.88,1,1}

\usepackage{multirow}
\usepackage{pgfplots}

\usepackage[nolist,nohyperlinks]{acronym}

\usepackage{subcaption}

\usepackage{xspace}
\usepackage{subfiles}
\usepackage{multirow}
\usepackage{tabularx}
\usepackage{subcaption}
\usepackage{enumitem}
\usepackage{balance}
\usepackage{xcolor}
\usepackage{cleveref}
\usepackage[splitrule,bottom]{footmisc}
\usepackage{longtable}
\usepackage{graphicx} 
\usepackage{rotating} 
\usepackage{lipsum}  
\usepackage{float} 
\usepackage{placeins} 
\usepackage{array} 
\usepackage{soul}
\usepackage{makecell}
\usepackage[font=small, skip=0.5ex]{caption}
\usepackage{amsmath}
\usepackage{booktabs} 

\usepackage{pifont}
\usepackage{longtable}
\usepackage{multirow}
\usepackage{array}

\newcommand{\sysname}{F\emph{in}P}

\pgfplotsset{compat=1.18}

\begin{document}

\begin{acronym}
    \acro{har}[HAR]{Human Activity Recognition}
    \acro{tcn}[TCN]{Temporal Convolutional Network }
    \acro{fl}[FL]{Federated Learning}
    \acro{aasd}[AASD]{Average Absolute SIA Difference From Mean}
    \acro{sia}[SIA]{Source Inference Attack}
    \acro{mia}[MIA]{Membership Inference Attack }
    \acro{mad}[MAD]{Mean Average Deviation}
    \acro{acc}[ACC]{Accuracy}
    \acro{rl}[RL]{Reinforcement learning}
    \acro{fedavg}[FedAvg]{Federated Averaging Algorithm}
    \acro{noniid}[non-IID]{Non-Independent and Identically Distributed}
    \acro{dp}[DF]{Differential Privacy}
    \acro{he}[HE]{Homomorphic Encryption}
    \acro{smpc}[SMPC]{Secure Multi-Party Computation}
    \acro{fe}[FE]{Functional Encryption}
\end{acronym}

\title[\sysname: Fairness-in-Privacy]{\sysname: Fairness-in-Privacy in Federated Learning by Addressing Disparities in Privacy Risk}

\author{Tianyu Zhao}
\orcid{0009-0000-3139-4872}
\affiliation{%
  \institution{University of California, Irvine}
  \city{Irvine}
  \state{California}
  \country{USA}}
\email{tianyu.zhao@uci.edu}

\author{Mahmoud Srewa}
\orcid{0009-0004-4795-4116}
\affiliation{%
  \institution{University of California, Irvine}
  \city{Irvine}
  \state{California}
  \country{USA}}
\email{msrewa@uci.edu}

\author{Salma Elmalaki}
\orcid{0000-0002-0969-0197}
\affiliation{%
\institution{University of California, Irvine}
  \city{Irvine}
  \state{California}
  \country{USA}}
\email{salma.elmalaki@uci.edu}

\begin{abstract}

Federated Learning (FL) inherently mitigates mass data centralization risks; however, its privacy protections are not equally distributed — leaving vulnerable individuals disproportionately exposed to sophisticated privacy attacks. Crucially, statistical heterogeneity in human-centric FL environments often results in an inequitable distribution of privacy risks, particularly affecting those whose sensitive attributes or behaviors make them outliers. To address this critical gap, we introduce \sysname, a novel framework designed to formalize and enforce fairness-in-privacy by mitigating disproportionate client vulnerability to Source Inference Attacks (SIA). \sysname\ operationalizes a two-pronged defense strategy that tackles both the symptoms and root causes of privacy disparity, ensuring that no group of clients bears an excessive privacy burden. It combines a server-side adaptive aggregation mechanism, which dynamically weights client contributions based on their estimated privacy risk, with a client-side regularization technique to curb localized overfitting that drives unique data memorization. Extensive empirical evaluations on FEMNIST, Human Activity Recognition (HAR), and CIFAR-10 datasets demonstrate that \sysname\ effectively aligns privacy fairness with primary task utility. Notably, \sysname\ successfully mitigates SIA risks and reduces disparities in privacy exposure, establishing that strong fairness-in-privacy guarantees need not compromise model utility. Ultimately, \sysname\ establishes equitable privacy protections by reducing vulnerability disparities by up to $57.14\%$, while preserving global model utility within a marginal $\pm 1.75\%$ of standard federated baselines.

\end{abstract}


\maketitle

\section{Introduction}\label{intro}

The pervasive integration of machine learning (ML) into human-centric applications demands careful consideration of both its ethical implications and its privacy guarantees. Historically, algorithmic fairness research has focused primarily on preventing bias and discrimination in model decisions or predictive accuracy~\cite{kleinberg2018algorithmic,rambachan2020economic, dwork2012fairness, kusner2017counterfactual, zhao2024fairo,zhao2024fina, elmalaki2021fair}. However, as ML systems increasingly rely on sensitive personal data, a critical yet underexplored dimension of fairness emerges: the fair distribution of privacy risks. Existing privacy definitions often focus on average or worst-case leakage across an entire dataset~\cite{mehrabi2021survey}. Yet, the crucial aspect of fair risk distribution—ensuring that no single individual or demographic group disproportionately shoulders the burden of privacy vulnerabilities—has received significantly less attention.

Federated Learning (FL) has emerged as an important and impactful privacy-enhancing technology to train ML models on decentralized devices. By enabling collaborative learning without centralizing raw data, FL inherently mitigates the risks of mass data collection and aligns well with stringent regulatory frameworks such as the GDPR and its Data Protection Impact Assessment (DPIA) provisions~\cite{GDPR, chang2024efficient, rieke2020future}. Despite these advantages, FL  remains vulnerable to sophisticated privacy attacks, including Membership Inference Attacks (MIA)~\cite{shokri2017membership} and Source Inference Attacks (SIA)~\cite{BG_SIA_2}, which aim to trace model memorization back to specific training data or participating clients. Crucially, the decentralized and statistically heterogeneous nature of FL — where data is not Independent and Identically Distributed (non-IID) — often amplifies unequal privacy leakage. When models overfit to clients with atypical or underrepresented data, those specific clients experience significantly higher privacy risks, creating a new form of digital inequality.

\paragraph{\textbf{Motivating Context and Societal Impact}}
The 2024 National Public Data (NPD) breach, which exposed billions of records, powerfully underscores the societal need for fairness in privacy~\cite{spectrumnews2024npdbreach}. Although the breach was widespread, its downstream consequences disproportionately impacted vulnerable populations—such as low-income individuals and the elderly—who have fewer resources to recover from identity theft and data abuse. Acknowledging the near inevitability of some data leakage—as reflected in the ``zero trust'' security paradigm~\cite{Rose2020Zero}—the security community's focus must expand from solely preventing breaches to ensuring equitable risk distribution when leaks occur. Within decentralized systems such as FL, this means formally defining and mitigating disproportionate harm. Formalizing the notion of fairness-in-privacy provides policymakers with the tangible metrics required to enforce equitable protections and fosters the public trust necessary for widespread adoption of AI.

\paragraph{\textbf{Disparate Risk in Human-Centric FL}}
To illustrate this disparity, consider the application of FL in Human Activity Recognition (HAR) for personalized health monitoring~\cite{rieke2020future,poulain2023improving}. In an FL-based HAR system, each user's wearable device acts as a client~\cite{BG_Survey2}. Users with physical disabilities or chronic conditions often exhibit movement patterns that deviate significantly from the ``average'' user in the global model. Consequently, the global model tends to memorize these unique features, making these marginalized individuals far more susceptible to re-identification via MIA~\cite{shokri2017membership} or SIA~\cite{BG_SIA_2}. This constitutes severe representational harm: individuals already facing societal vulnerabilities are disproportionately exposed to privacy risks simply because their data is statistically unique. Without explicitly addressing this inequitable distribution of risk, FL risks reinforcing the very social inequalities it should theoretically bypass.

\paragraph{\textbf{Proposed Approach}}
This paper addresses the critical challenge of ensuring equitable privacy protection by introducing \sysname, a framework designed to measure and enforce fairness in privacy within FL. To mitigate disparate privacy leakage caused by heterogeneous data and varying local training dynamics~\cite{selialia2024mitigating}, we propose a novel two-pronged approach operating at both the server and the client levels:

\begin{enumerate}
\item \textbf{Server-Side Adaptive Aggregation:} We introduce an aggregation strategy that dynamically weights client model updates based on their estimated privacy risk. 
This prevents highly vulnerable, unique clients from disproportionately exposing their representations in the global model.
\item \textbf{Client-Side Collaborative Overfitting Reduction:} Complementing the server-side intervention, we tackle the root cause of vulnerability—local overfitting. By estimating each client's relative overfitting using the maximum eigenvalue of its local model's Hessian, we incorporate this metric into a local regularization term bounded by the Lipschitz constant. This encourages clients to learn generalizable representations, directly reducing their individual susceptibility to privacy attacks.
\end{enumerate}

By addressing both the symptoms (unequal risk aggregation) and the root causes (local overfitting) of privacy disparity, our combined approach achieves an improvement in fairness in privacy with a negligible impact on primary task performance.

\paragraph{\textbf{Contributions}}
In summary, our key contributions are:
\begin{itemize}
\item We introduce a formal definition of fairness-in-privacy (\sysname) tailored for real-world human-centric machine learning systems.
\item We operationalize the measurement of \sysname\ within Federated Learning, with a specific focus on quantifying vulnerabilities to source inference attacks (SIA).
\item We propose a novel, two-pronged technical framework featuring adaptive server-side aggregation and client-side Hessian-based regularization to optimize \sysname.
\item We provide a comprehensive empirical evaluation demonstrating the efficacy of our approach using real-world dataset of a human activity recognition (HAR), and standard learning tasks (FEMNIST and CIFAR-10), confirming significant gains in fairness in privacy with minimal task utility loss.
\end{itemize}

\section{Background and Related Work}\label{sec:related}

\paragraph{\textbf{Privacy in Human-Centric Systems}}
Ensuring privacy in human-centric machine learning systems presents inherent conflicts among service utility, computational cost, and personal privacy~\cite{sztipanovits2019science}. Without appropriate frameworks and incentives for secure data utilization, system deployments often face a dichotomy: policies become too restrictive, stifling innovation, or compromise private information, leading to adverse selection and a loss of public trust~\cite{jin2016enabling, mulligan2016privacy, fox2021exploring, goldfarb2012shifts}. Consequently, extensive research has focused on establishing privacy-preserving techniques and auditing platforms for human-in-the-loop and decision-making systems~\cite{abadi2016deep, cummings2019compatibility, taherisadr2023adaparl, taherisadr2024hilt, jagielski2020auditing, raji2020saving, elmalaki2022vindico}. However, recognizing the ``zero trust'' reality that perfect privacy is often mathematically or practically unattainable, this paper examines privacy through an equity lens. We investigate how to ensure a fair distribution of harm when privacy leakage inevitably occurs, bridging the technical challenges of ML with the ethical imperatives of equitable protection.

\paragraph{\textbf{Privacy Disparity and  Inference Attacks in Federated Learning.} }
\acf{fl} enables the collaborative training of models without centralizing raw data~\cite{mcmahan2017communication}, making it highly suitable for privacy-sensitive domains navigating regulations, such as GDPR~\cite{BG_Survey2, BG_Survey1}. Despite its decentralized architecture, FL introduces new attack surfaces. The statistical heterogeneity inherent in FL does not only affect model performance — it systematically amplifies the disparity in privacy risk between participating clients. When client data distributions diverge significantly, models tend to memorize the unique features of outlier clients, making those clients disproportionately vulnerable to inference attacks. Foundational work on disparate vulnerability~\cite{kulynych2022disparate} has demonstrated that privacy leakage is rarely uniform; models disproportionately memorize and expose the data of underrepresented subgroups. Membership Inference Attacks (MIA) exploit this memorization by determining whether a specific data record was used during training, effectively probing the model's generalization boundary~\cite{shokri2017membership, BG_MIA}. Although MIA captures global membership leakage, it does not reveal which client contributed to the targeted record~\cite{BG_MIA_1, BG_MIA_2}. Source Inference Attacks (SIA) extend this threat by identifying the exact originating client of a training record, directly exploiting localized overfitting caused by non-IID data distributions~\cite{BG_SIA_2}. Despite pseudonymity, SIA poses a serious risk because mapping sensitive data to a distinct client node links it to a real-world context such as a device or location, effectively breaking anonymity. Beyond membership-based attacks, Property Inference Attacks (PIA) represent a distinct threat class that targets macro-level statistical properties of a client's local dataset — such as demographic distributions or class ratios — rather than individual records~\cite{melis2019exploiting}. Unlike SIA and MIA, which are rooted in localized memorization and overfitting, PIA exploit the structural properties embedded in gradient updates during the aggregation process and, therefore, operate from a fundamentally different threat vantage point. 

Crucially, the inherent statistical heterogeneity (\ac{noniid} data) of human-centric FL amplifies these vulnerabilities. When data distributions differ widely among clients, individual model updates become highly distinct. Malicious actors can exploit these distributional differences to trace memorized features to specific clients~\cite{BG_NON_IID}. This distinctiveness is especially pronounced in datasets like \ac{har}, where unique physical movements or conditions make outliers highly susceptible to SIAs, creating an unequal landscape of privacy risk.

Existing defenses such as Differential Privacy via DP-SGD~\cite{abadi2016deep} address these risks by injecting uniform noise into gradient updates. However, as demonstrated in this work, geometry-blind noise injection degrades global utility without resolving the disparity in per-client privacy exposure; in fact, it can exacerbate unfairness by failing to mask the highly distinct updates of severely skewed clients while unnecessarily penalizing well-represented ones.

\paragraph{\textbf{From Utility Fairness to Privacy Fairness.}} Fairness-aware federated learning approaches have emerged to address performance inequity between clients, including techniques such as FedAvg~\cite{hu2023source} with client reweighting, FairFed ~\cite{BG_Fairness_2}, and personalized federated learning methods~\cite{BG_Personalization,BG_Personalization_2}, as well as recent work exploring the tension between fairness interventions and privacy leakage~\cite{bendoukha2025towards}. Recent work also investigated the use of FL for pluralistic alignment of Large Language Models with a fairness lens of aggregation of human preference~\cite{srewa2026appa, srewa2025pluralllm, srewa2025systematic}.

However, these approaches primarily frame fairness in terms of equitable model accuracy or predictive benefit across client groups, and do not explicitly formalize or optimize for the equitable distribution of privacy risk. The critical dimension of client-level privacy equity — ensuring that no individual client disproportionately bears the burden of privacy vulnerability due to the statistical uniqueness of their data — remains largely unaddressed in the prior literature. \sysname uniquely targets this gap by formalizing fairness-in-privacy as a first-class objective, directly mitigating memorization-based privacy disparities caused by non-IID data distributions \textbf{exploited by SIA}, and distinguishing itself from existing defenses that focus either on global privacy guarantees or on utility fairness without addressing the structural inequity of per-client privacy exposure.

\section{Threat Model and Problem Formulation}\label{sec:threatmodel}

Although Federated Learning (FL) mitigates the risks of centralized data storage, it remains vulnerable to inference attacks launched by the aggregating server.

\paragraph{\textbf{Assumptions and Scope}} \sysname\ operates under two core assumptions. First, the server is ``honest-but-curious'': it faithfully executes the FL protocol but actively attempts to infer sensitive client information from received model updates and curvature metrics. Second, clients are honest and cooperative: they follow the protocol and submit genuine updates without manipulation. Under these assumptions, \sysname\ targets inference attacks driven by localized overfitting and memorization in non-IID settings — specifically Source Inference Attacks (SIA), which identify the originating client. SIA is particularly risky because mapping a sensitive record to a specific client node links it to a real-world context such as a device or location, breaking pseudonymity through contextual profiling. \sysname\ suppresses this memorization by flattening the loss landscape, reducing the attacker's confidence. \sysname\ does not address malicious server behavior, manipulative clients, gradient reconstruction attacks, or Property Inference Attacks, as these operate outside the memorization-based threat model.

\paragraph{\textbf{Threat Model}} A primary privacy threat in this decentralized setting is a two-stage inference attack executed by the server: a Membership Inference Attack (MIA) followed by a Source Inference Attack (SIA).

\begin{itemize}[noitemsep, topsep=0pt]
    \item \textbf{Membership Inference Attack (MIA):} The server first attempts to determine if a specific target data point $x$ was utilized in the training set $D_{\theta_g}$ of the global model $\theta_g$. This is formally defined as the probability that $x$ belongs to the training data:
    $$MIA(\theta_g, x) = P(x \in D_{\theta_g})$$
    
    \item \textbf{Source Inference Attack (SIA):} If the MIA successfully indicates that $x$ was part of the training corpus, the server proceeds to identify the origin of the data. The server analyzes the updates of the local model $\theta_i$ to determine the probability that a specific client $C_i$ is the source of the record $x$:
    $$SIA(\theta_i, x) = P(C_i \mid x, \theta_i)$$
\end{itemize}

As demonstrated in the prior literature~\cite{BG_SIA_2}, the concatenation of these attacks compromises the anonymity of the client. Furthermore, auditing and completely preventing MIA and SIA within FL presents inherent mathematical and practical limitations~\cite{chang2024efficient}.

\paragraph{\textbf{Problem Formulation: Disparate Privacy Risk}}
Rather than attempting the mathematically fraught task of completely eliminating SIA, our work focuses on the resulting \textit{disparity} in privacy risk across the client federation. In heterogeneous (non-IID) FL environments, clients with unique or outlier data distributions are highly prone to local overfitting during training. This overfitting causes their local model updates ($\theta_i$) to disproportionately memorize their unique data points, rendering them significantly more vulnerable to SIAs than clients with more ``average'' data. To exploit this vulnerability, The attacker calculates the prediction loss of the target record across every individual client's model on the server. The reason is that the local model belonging to the true source will naturally yield the lowest prediction loss. The attacker therefore simply identifies the client with the minimum loss as the most probable source of the target record. This creates a systemic inequity in privacy protection. 

Given this threat model, our objective is to mitigate the disparate impact of SIAs by ensuring an equitable distribution of the inherent privacy risks between all clients. To achieve this \sysname, our framework pursues two core objectives:

\begin{enumerate}[noitemsep, topsep=0pt, start=1, label={(\bfseries O\arabic*):}]
    \item \textbf{Addressing the Symptoms (Server-Side):} Develop a dynamic aggregation methodology on the server that quantifies client vulnerability and adjusts their contributions to the global model, ensuring a fair distribution of privacy risk without degrading global utility.
    \item \textbf{Addressing the Root Causes (Client-Side):} Provide actionable feedback mechanisms to highly vulnerable clients, enabling them to apply targeted local regularization. This empowers clients to reduce their local overfitting, thereby lowering their individual susceptibility to SIAs and improving the overall fairness in privacy of the system.
\end{enumerate}

\section{The \sysname\ Framework in Federated Learning}\label{sec:finpmetric}

\begin{figure*}[!t]
\centering
\includegraphics[trim={0 9cm 3cm 0},clip,width=0.8\linewidth]{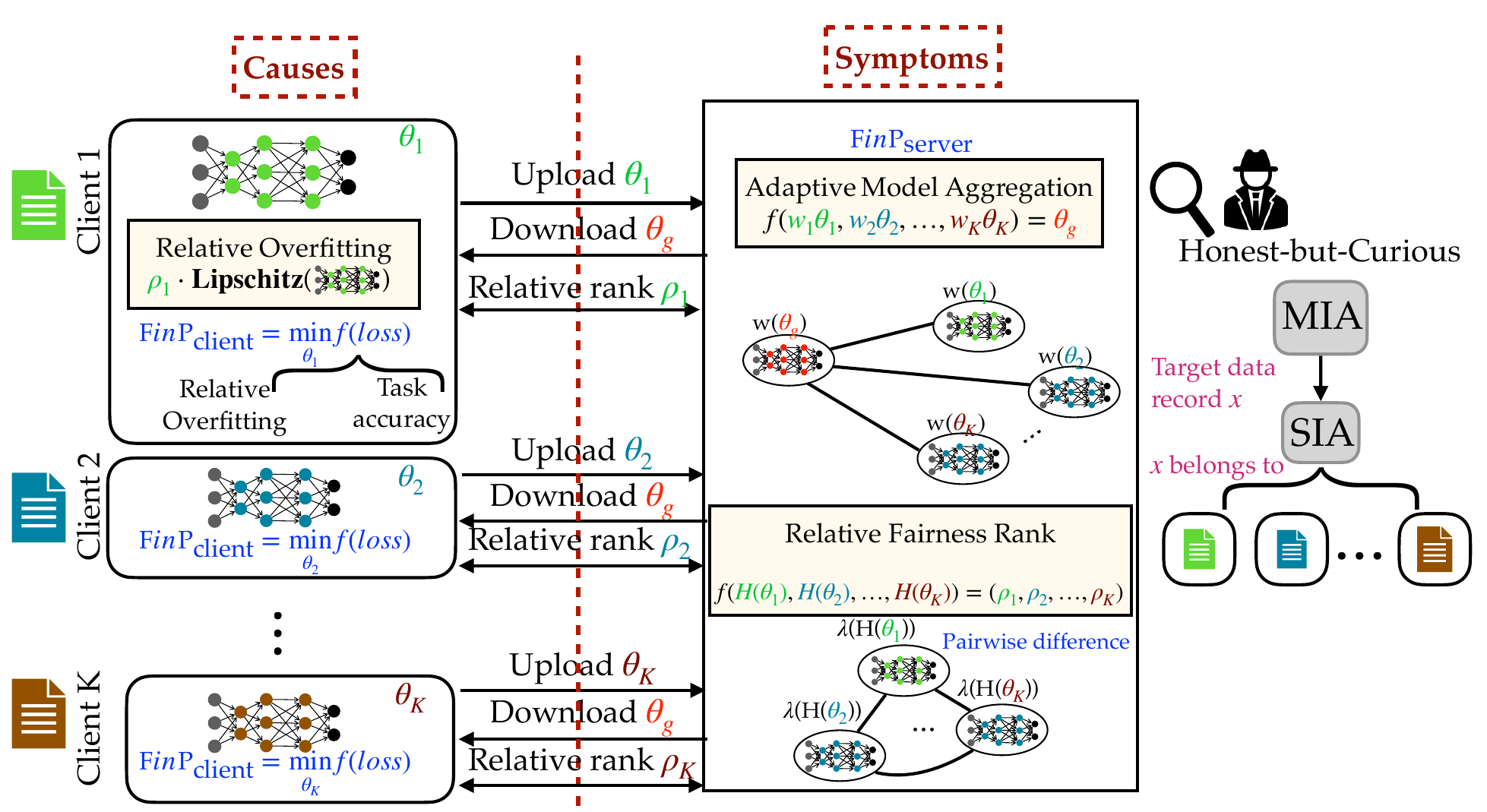}
\caption{Overview of the \sysname\ framework. The system achieves fairness-in-privacy by establishing a feedback loop that simultaneously addresses the symptoms of privacy disparity (via server-side adaptive aggregation) and the root causes (via client-side collaborative overfitting reduction).}
\label{fig:framework}
\end{figure*}

The core objective of the \sysname\ framework is to ensure that privacy risks are equitably distributed among all participating clients, preventing any single client from bearing a disproportionate burden of vulnerability. In practical federated learning (FL) deployments, the profound heterogeneity of client data distributions, varying computational resources, and distinct local training dynamics naturally lead to stark disparities in privacy risk. Traditional privacy defenses that merely attempt to lower the \textit{average} risk across the federation are insufficient for human-centric systems; effective protection demands that we proactively prevent disproportionate risk burdens on minority or outlier clients.

To achieve true fairness-in-privacy, \sysname\ operationalizes a holistic two-pronged approach, as illustrated in \autoref{fig:framework}. We argue that systemic privacy equity requires simultaneous intervention at both the server level (during model aggregation) and the client level (during local training). 

\paragraph{\textbf{Addressing the Symptoms (Server-Side Intervention)}}
Server-side interventions are critical for managing the immediate impacts of existing privacy disparities. \sysname\ introduces an adaptive aggregation mechanism that dynamically scales the weights of client updates based on their estimated privacy risk. By systematically reducing the aggregation weight of highly vulnerable clients, we prevent their overfit, memorized representations from unduly influencing the global model. This limits the exposure of their unique data to the rest of the federation and restricts an adversary's ability to execute Source Inference Attacks (SIAs). However, while adaptive aggregation effectively mitigates the immediate symptoms of privacy unfairness, it does not cure the underlying vulnerability of clients themselves. 

\paragraph{\textbf{Addressing the Root Causes (Client-Side Intervention)}}
To dismantle the root cause of privacy disparity, we must look to local training dynamics---specifically, local overfitting. Extensive prior literature has established the theoretical and empirical connection between model overfitting and increased privacy risks~\cite{BG_SIA_2, Yeomoverfitting, shokri2017membership}. Because overfit models memorize specific properties of their training data rather than learning generalized features, they serve as the primary vulnerability exploited by MIAs and SIAs. Consequently, when a client's local model heavily overfits its non-IID data, that client's susceptibility to SIAs spikes, creating the very disparity we aim to eliminate.

To counteract this, \sysname\ introduces a \textbf{collaborative overfitting reduction strategy} at the client level. This proactive mechanism ranks clients based on an estimation of their relative overfitting compared to the federation. By incorporating this server-provided rank into a customized local regularization scheme, \sysname\ forces vulnerable clients to learn more generalizable representations. This actively suppresses the memorization of outlier data, shrinking the initial disparity in privacy risk \textit{before} the models are ever transmitted to the server. 

\paragraph{\textbf{The Closed-Loop System}}
Ultimately, \sysname\ functions as a tightly coupled closed-loop system. The server utilizes incoming client models to calculate adaptive aggregation weights and relative overfitting ranks; in turn, the clients utilize this targeted feedback from the server to regularize their subsequent local training rounds. By simultaneously addressing both the symptoms and the root causes of privacy disparity, \sysname\ establishes a structurally fair FL environment.

We formally define the mathematical mechanics of the client component in \autoref{sec:client_side_adjustments} and the server component in \autoref{sec:server_side_adjustment}.

\subsection{Addressing the Root Causes: Client-Side Collaborative Regularization}\label{sec:client_side_adjustments}

Although server-side aggregation mitigates the immediate symptoms of privacy unfairness, achieving structural fairness-in-privacy (\sysname) requires dismantling the root cause: local overfitting. In a heterogeneous FL system with $K$ clients, non-IID data create distributional shifts that cause certain outlier clients to overfit more than others. Because overfitting directly correlates with the memorization of training data, these specific clients become disproportionately susceptible to Source Inference Attacks (SIAs). Our goal is to enforce an equitable privacy risk by proactively reducing this localized memorization.

To achieve this, we introduce a collaborative, feedback-driven regularization strategy. Recent literature demonstrates that the top Hessian eigenvalue ($\lambda_{\text{max}}$) and the Hessian trace ($H_T$) are powerful metrics to characterize the loss landscape and the generalization capabilities of neural networks~\cite{Jiang2020Fantastic}. A sharp loss landscape (indicated by high $\lambda_{\text{max}}$ and $H_T$) implies severe overfitting and a high vulnerability to weight perturbations, while lower values indicate smoother, more generalized and thus more privacy-preserving local models.

Because we are specifically interested in \sysname---which focuses on \textit{disparity} rather than absolute risk---we quantify how much more vulnerable a client is compared to its peers. We define each client's relative overfitting by calculating the average pairwise difference across the top Hessian eigenvalue ($\lambda_{\text{max}}^k$) and Hessian trace ($H_T^k$) of their local model $\theta_k$ against all other $j$ clients:

\begin{align} 
\bar{\Delta}_k&=\frac{1}{K-1}\sum_{j=1, j\neq k}^{K}|\lambda_{\text{max}}^k-\lambda_{\text{max}}^j| \\
\bar{H}_k&=\frac{1}{K-1}\sum_{j=1, j\neq k}^{K}|H_T^k-H_T^j| 
\label{eq:hessian}
\end{align}

These metrics are then normalized and combined to compute the client's \textit{Overfitting Relative Rank} ($\rho_k$), which serves as a proxy for their relative privacy risk disparity:

\begin{align}\label{eq:overfittingrank}
\rho_k=\frac{1}{2}\left(\frac{\bar{\Delta}_k}{\max(\bar{\Delta})}+\frac{\bar{H}_k}{\max(\bar{H})}\right) 
\end{align}

Clients compute their own top Hessian eigenvalue and Hessian trace locally during training. The server only computes the relative rank $\rho_k$ after receiving these scalar metrics. The server then transmits this scalar rank $\rho_k$ to client $k$ as a targeted feedback.

Upon receiving $\rho_k$, the client incorporates this overfitting rank into their subsequent local training rounds using a dynamic regularization term. We utilize the Lipschitz constant, approximated by the spectral norm of the Jacobian matrix ($||J_k||$)~\cite{liu2020simple}. Constraining the Lipschitz constant forces the model to learn smoother functions, thereby resisting the memorization of outlier data points. The modified local loss function for client $k$ becomes:
\begin{align} 
\mathcal{L}_k^*&=\mathcal{L}_k+\beta\cdot\rho_k\cdot||J_k|| \label{eq:lossclient}\\ 
\text{\sysname}_{\text{client}}&=\min_{\theta_k}\mathcal{L}_k^* 
\label{eq:finpclient}
\end{align}

Here, $\mathcal{L}_k$ is the original local loss function, $||J_k||$ is the Jacobian penalty, and $\beta$ is a task-dependent parameter that adjusts the baseline impact of Lipschitz regularization. The critical innovation is the inclusion of $\rho_k$, which acts as an \textbf{adaptive} multiplier. In each communication round, clients with a higher relative overfitting rank are subjected to mathematically stronger regularization. This collaborative, penalty-based approach effectively throttles the most vulnerable clients, forcing them to generalize, and thereby actively shrinking the privacy risk disparity across the federation.

\paragraph{\textbf{Computational Practicality and Privacy Guarantee}}
Computing and transmitting the exact Hessian matrix is computationally prohibitive for deep neural networks and violates the strict data-minimization guarantee of FL. To ensure that our framework remains scalable, clients do not compute the full Hessian; instead, they efficiently estimate only the top eigenvalue and the trace. 

To minimize local computational overhead on edge devices — including battery-constrained mobile phones and resource-limited edge nodes — clients employ efficient Hessian-vector product (HVP) methods, specifically power iteration for the top eigenvalue and Hutchinson's method for the trace, entirely avoiding full Hessian instantiation. Critically, because \sysname's fairness objective relies solely on the relative ranking of client vulnerability rather than the absolute values of eigenvalues, these lightweight approximations are fully sufficient to achieve privacy equity without requiring exact Hessian computation. This design choice substantially reduces computational cost per-round, making \sysname\ practical for deployment on resource-constrained hardware typical in the real-world federated participants.

\sysname\ does not require additional communication rounds to transmit these metrics. The scalar sharpness values are simply appended to and transmitted concurrently with the standard local model weights during existing federated communication rounds, incurring only negligible additional bandwidth overhead and fully preserving the communication efficiency of standard federated learning.

Furthermore, while \sysname\ requires clients to transmit these local curvature metrics with their model updates($\theta_k$), this does not violate the FL privacy principles. These metrics are one-dimensional scalars representing macroscopic geometric properties of the local loss landscape. Because mapping a local dataset to these two aggregate scalars is a heavily non-injective and non-invertible transformation, transmitting them provides negligible mutual information regarding the raw training data. It is mathematically intractable for an honest-but-curious server to reconstruct raw local data or execute gradient-inversion attacks relying solely on these curvature scalars.

\subsection{Addressing the Symptoms: Server-Side Adaptive Aggregation}\label{sec:server_side_adjustment}

The core symptom of privacy unfairness in FL is the pronounced disparity in vulnerability to Source Inference Attacks (SIAs) across the client federation. While client-side regularization (detailed in Section~\ref{sec:client_side_adjustments}) addresses the root cause of local overfitting, the server must simultaneously manage the observable symptoms present in the incoming model updates. Our objective is to design a server-side aggregation strategy that actively quantifies this privacy risk and adjusts the global model update to enforce an equitable distribution of vulnerability. We explicitly define this objective as minimizing the variance of the privacy risk distribution:

\begin{align}\label{eq:finpserver}
\begin{split}
    \text{\sysname}_{\text{server}} &= \min_{\mathbf{w} \in \mathcal{W}} \text{Disparity}(\mathbf{p}(\mathbf{w})) \\
    &= \min_{\mathbf{w} \in \mathcal{W}} \left\| \mathbf{p}(\mathbf{w}) - \frac{1}{K} \mathds{1}^T \mathbf{p}(\mathbf{w}) \otimes \mathds{1} \right\| + \left\|\frac{1}{K} \mathds{1}^T \mathbf{p}(\mathbf{w})\right\|
\end{split}
\end{align}

Here, $\mathbf{p}(\mathbf{w})) = [p_1(\mathbf{w})), \dots, p_K(\mathbf{w}))]$ is the vector of privacy risk indicators across the $K$ clients given the aggregation weights $\mathbf{w}$, $\mathds{1}$ is a vector of ones of length $K$, and $\mathcal{W} = \{\mathbf{w} \in \mathbb{R}^K \mid \sum_{k=1}^{K} w_k = 1, w_k \geq 0 \ \forall k\}$ is the set of valid aggregation weights. 

To satisfy the $\text{\sysname}_{\text{server}}$ objective, the server must aggregate local models in a manner that minimizes this disparity metric without severely degrading global utility. We evaluate two strategies for achieving this:

\begin{enumerate}[leftmargin=*]

    \item \textbf{Strategy 1: Adaptive Lightweight Aggregation (ALA) (Proposed Solution)}\\
    Our primary method acts as a highly efficient, heuristic solution to the disparity minimization objective, specifically designed to be scalable for real-world, resource-constrained environments.
    \begin{itemize}
        \item \textbf{Risk Symptom Quantification ($p_k$):} The server re-utilizes the normalized Overfitting Relative Rank ($\rho_k$) described in Equation~\ref{eq:overfittingrank}, as a proxy for the privacy risk symptom ($p_k \propto \rho_k$). A higher $\rho_k$ signifies severe local overfitting and greater vulnerability to SIAs.
        \item \textbf{The Solution Mechanism:} The ALA technique achieves the disparity minimization objective by dynamically calculating the specific aggregation weights $\alpha_k \in \mathcal{W}$ to be inversely proportional to the risk symptom $\rho_k$. High-risk clients are explicitly granted less influence over the global model, proactively driving the global model toward a state where the risk vector $\mathbf{p}$ is balanced:

        \begin{align}\label{eq:ala}
            \mathbf{\theta}_{global} \leftarrow \sum_{k=1}^{K} \alpha_k \mathbf{\theta}_k, \quad \text{where} \quad \alpha_k = \frac{1 - \rho_k}{\sum_{i=1}^{K} (1 - \rho_i)},  
        \end{align}

        where $\theta_{global}$ is the global model parameters and $\theta_{k}$ is the local model parameters for client $k$. This inverse weighting mechanism provides the exact control necessary to solve Equation~\ref{eq:finpserver}. By actively restricting the impact of the most vulnerable clients, ALA prevents their memorized, outlier data from being embedded into the global model. This enforces an equitable distribution of privacy risk without introducing significant computational overhead.
    \end{itemize}

    \item \textbf{Strategy 2: Server-Side Risk Quantification (PCA-based Baseline)}\\
    For comparison only, a traditional alternative for quantifying the symptom of risk $p_k$ is computing the Principal Component Analysis (PCA) distance between each client's model update and the global average update~\cite{durmus2021federated}. This distance acts as a purely server-side proxy for outlier behavior. 

    If this PCA distance ($PCA_d$) is used as the risk vector $\mathbf{p}(\mathbf{w})$, the aggregation process would similarly solve the $\text{FinP}_{\text{server}}$ disparity minimization objective by adjusting weights based on these distances. However, while formally sound, computing the PCA distance for the high-dimensional model updates typical in modern human-centric FL is computationally expensive~\cite{candes2011robust}. The resulting latency makes it highly impractical for real-world, large-scale FL deployments, reinforcing that the lightweight, Hessian-proxy ALA strategy is the superior and preferred mitigation method for our framework.

\end{enumerate}

\section{Evaluation}\label{sec:eval}

\subsection{Federated Learning System Setup}\label{sec:evalsetup}

To demonstrate the real-world applicability of \sysname\ to mitigate privacy disparities, we evaluated our framework in two distinct federated learning deployments. We specifically selected datasets that reflect the highly heterogeneous, non-IID nature of practical edge-computing and mobile sensing environments, where outlier clients are naturally prone to local overfitting and heightened vulnerability to Source Inference Attacks (SIAs). In particular, we structured our evaluation around two primary case studies: Human Activity Recognition (HAR) as a human-centric, privacy-sensitive application, and CIFAR-10 and FEMNIST as a standard benchmark for controlled, heterogeneous visual data. 

\paragraph{\textbf{Baselines and Ablations}}
To rigorously isolate the contributions of our framework, we compare the three standard and state-of-the-art baselines in various configurations as follows: 

$\bullet$ \textbf{HAR dataset:} we evaluated five configurations:
(1) Baseline FL: Standard FedAvg aggregation, adapted from prior SIA literature~\cite{hu2023source}. (2) \sysname$_\text{server}$: Isolating our adaptive server-side aggregation without client collaboration (Equation~\ref{eq:finpserver}). (3) \sysname$_\text{client}$: Isolating our collaborative client-side regularization without adaptive server aggregation. (4) \sysname-Full: The complete proposed closed-loop framework. (5) Differential Privacy (DP-SGD)~\cite{abadi2016deep}: Applying DP-SGD against SIA attack with noise multiplier $\sigma_{DP}\in \{0.5, 1, 2\}$.

$\bullet$ \textbf{CIFAR-10 dataset:} we evaluated:
(1) Baseline FL: Standard FedAvg from~\cite{hu2023source}. (2) FedAlign: A state-of-the-art FL method specifically designed to address non-IID data heterogeneity~\citep{mendieta2022local}. (3) \sysname\ (ResNet56): Our framework applied to the same ResNet architecture used by FedAlign for a direct performance comparison. (4) \sysname\ Ablation (CNN): Our framework applied to a standard CNN to evaluate the sensitivity of the impact factor $\beta \in \{0.05, 0.1, 0.3, 0.5\}$, as described in Equation~\ref{eq:lossclient}. (5) Differential Privacy (DP-SGD)~\cite{abadi2016deep}: Applying DP-SGD against SIA attack with noise multiplier $\sigma_{DP}\in \{0.5, 1, 2\}$.
(6) Additionally, we evaluate \sysname\ using both the Adaptive Lightweight Aggregation (ALA) and the PCA-based server-side strategies to validate computational practicality.

$\bullet$ \textbf{FEMNIST dataset:} we evaluate:
(1) Baseline FL: Standard FedAvg from~\cite{hu2023source}. (2) \sysname: Our framework with the impact factor $\beta \in \{0.5, 0.75, 1\}$, as described in Equation~\ref{eq:lossclient}. (3) Differential Privacy (DP-SGD)~\cite{abadi2016deep}: Applying DP-SGD against SIA attack with noise multiplier $\sigma_{DP}\in \{0.5, 1, 2\}$. (4) MIA attack: Launching White-Box Membership Inference Attack (MIA) to evaluate \sysname performance vs. Baseline (FedAvg) and DP-SGD. 

\paragraph{\textbf{Setup for HAR}} 
The UCI HAR Dataset ~\cite{human_activity_recognition_using_smartphones_240} serves as our  real-world application. In modern ubiquitous computing and mobile sensing environments, continuous sensor data (such as accelerometer and gyroscope readings) is routinely collected from users' personal smartphones or wearable devices. Because physical behaviors, daily routines, and gaits are unique to an individual, HAR data is inherently non-IID and deeply sensitive~\citep{har2,har3}. Preserving the natural user-based partition ensures the authentic structure of the data is maintained; models trained on users with distinct or outlier behaviors are prime targets for SIAs.

The HAR dataset comprises data from 30 subjects (aged 19--48), treated as $K=30$ individual FL clients, performing six daily activities. We allocated $70\%$ of each client's data for training (using 5-fold cross-validation) and $30\%$ for independent testing. All data was preprocessed with noise filters and segmented using a 2.56-second sliding window with a $50\%$ overlap. We utilized a Temporal Convolutional Network (TCN)~\cite{bai2018empirical} for the architecture of the local models, which captures temporal dependencies via causal convolutions. We trained over 20 global communication rounds using FedAvg. Clients trained locally with a batch size of $64$, a learning rate of $0.001$ (Adam optimizer), 1 local epoch per round, and an impact factor $\beta = 2$. More details on the HAR dataset are in Appendix~\ref{appendix:dataset}.

To establish a rigorous Differential Privacy (DP) baseline, we implemented Record-Level DP-SGD~\cite{abadi2016deep} utilizing the Opacus library. For all DP-enabled experiments, we transitioned from the Adam optimizer to Stochastic Gradient Descent (SGD) with a learning rate of 0.01, while keeping all other federated configurations (batch size, local epochs, and communication rounds, etc) identical to the standard baseline. A critical challenge in DP-SGD is selecting an optimal clipping threshold (C) that aggressively bounds the sensitivity of extreme outliers while preserving the underlying gradient signal. To ensure an empirical and fair calibration, we executed a non-private warm-up run of the global model for a single epoch and recorded the unclipped $L_2$ gradient norms across all client batches. We fixed the clipping threshold at C=5, which closely approximates the 90th percentile of the unclipped gradients. By adopting this empirically derived threshold, we formally bounded the model's sensitivity while maintaining optimal learning momentum. To comprehensively evaluate the resulting privacy-utility trade-off, we swept the Gaussian noise multiplier $\sigma_{DP}\in \{0.5, 1, 2\}$, corresponding to low, medium, and high privacy budgets, respectively.

\begin{figure}[!t]
\centering
\includegraphics[width=\linewidth]{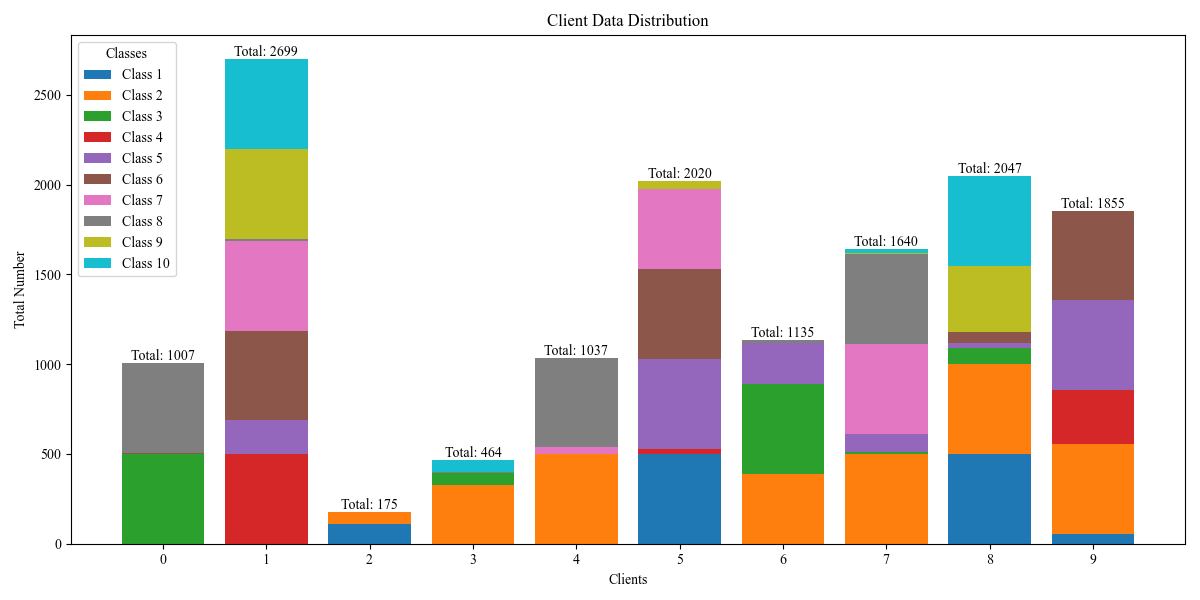}
\caption{CIFAR dataset profile demonstrating severe non-IID data distribution among clients after Dirichlet sampling ($\alpha=0.1$).}
\label{fig:CIFAR data}
\vspace{-3mm}
\end{figure}

\paragraph{\textbf{Setup for CIFAR-10}}
To stress-test \sysname\ under mathematically controlled degrees of data imbalance---simulating visual heterogeneity across a network of personal devices or edge cameras---we utilized the CIFAR-10 dataset~\citep{krizhevsky09learning}. We partitioned the data among $K=10$ clients using a Dirichlet distribution $Dir(\alpha)$ with $\alpha=0.5$, a standard approach in SIA literature for simulating unbalanced subsets~\citep{mendieta2022local,BG_SIA_2}. \autoref{fig:CIFAR data} visualizes the severe data skew among clients when tested at $\alpha=0.1$. 
We employed ResNet56~\citep{he2016deep} for the primary evaluation to match the FedAlign baseline, and a standard CNN~\cite{hu2023source} for the $\beta$ ablation studies. Similar to the HAR setup, training occurred over 20 global rounds. The default impact factor for the client-side regularization was set to $\beta = 0.3$. We deployed DP-SGD with CNN and same configuration. Clipping threshold is fixed at $C=1.75$ ($\approx$ the 90th percentile of the unclipped gradients) and noise multiplier $\sigma_{DP}\in \{0.5, 1, 2\}$ for low, medium, high privacy budgets.
More details on the CIFAR dataset are in Appendix~\ref{appendix:dataset}.

\paragraph{\textbf{Setup for FEMNIST}}
We evaluate on FEMNIST Dataset additionally, a handwritten character recognition benchmark with 62 classes and natural non-IID structure induced by writer identity. We simulate $10$ federated clients by sampling $10$ distinct writers; each client owns only that writer’s examples. We employ a lightweight CNN as the training model. More details are in Appendix \ref{appendix:dataset}.

\paragraph{\textbf{Empirical Evaluation Setup}}
To simulate this attack pipeline for our empirical evaluation, we randomly sampled training data from each client's local dataset and pooled them into a unified dataset of target records. By launching the SIA against these known records and calculating the attack success rate, we can directly measure both the absolute privacy leakage and, crucially, the \textit{disparity} of this vulnerability across the federation.

\paragraph{\textbf{Hardware Profile}} 
All experiments were conducted on a single NVIDIA A30 GPU with 24 GB of memory, simulating the central aggregating server's computational environment.

\subsection{Metrics for Comparison}\label{sec:metrics}

To comprehensively evaluate the effectiveness of \sysname\ in achieving fairness in privacy, we employ the following key metrics encompassing privacy fairness, absolute privacy, utility, and efficiency. 

\paragraph{\textbf{1. Privacy Fairness Metrics (Disparity Quantification)}}
To quantify ``Fairness in Privacy,'' we must measure the dispersion of Source Inference Attack (SIA) vulnerability across the $K$ participating clients. We introduce three metrics to capture both the accuracy and the confidence of the adversary's attack.

\begin{itemize}[leftmargin=*]
    \item \textbf{Disparity in SIA Accuracy (CoV \& FI):} 
To formalize the notion of privacy fairness, we draw on the concept of group fairness as established in literature~\cite{jain1984quantitative}. Applying the Coefficient of Variation (CoV) to client-level privacy risk treats each client as a protected group, where a low CoV indicates an equitable distribution of privacy exposure across the federation. 
Unlike minimax or worst-case approaches that optimize solely for the single most vulnerable client, minimizing CoV ensures system-wide privacy equity, preventing minority clients from suffering disproportionately high privacy vulnerability while preserving equitable protection across the entire federation.

For a federation of $K$ clients, the empirical SIA accuracy for a specific client $i$ is calculated as:
    \begin{align}
        \text{SIA}_{acc_i} &= \frac{\text{Number of correct SIA identifications for client } i}{\text{Total target records attributed to client } i}
        \label{eq:accsia}
    \end{align}
    Let $\mu = \frac{1}{K} \sum_{i=1}^{K} \text{SIA}_{acc_i}$ be the mean SIA accuracy across the federation, and $\sigma$ be the standard deviation. The CoV is computed as:
    \begin{align}
        \text{CoV}(\text{SIA}_{acc}) &= \frac{\sigma}{\mu} = \frac{\sqrt{\frac{1}{K} \sum_{i=1}^{K} (\text{SIA}_{acc_i} - \mu)^2}}{\mu} 
        \label{eq:covsia}
    \end{align}
    To map this variance to an intuitive fairness percentage between 0 and 1 (where 1 represents perfect equity), we apply the Fairness Index (FI) transformation:
    \begin{align}
        \text{FI}(\text{SIA}_{acc}) &= \frac{1}{1 + \text{CoV}(\text{SIA}_{acc})^2} 
        \label{eq:fisia}
    \end{align}
    A higher FI value indicates that the vulnerability is equitably distributed, preventing any single user from becoming an extreme outlier target.

    \item \textbf{Equal Opportunity Difference in Privacy (EOD):} Because $\text{SIA}_{acc_i}$ represents the empirical True Positive Rate (TPR) of the adversary correctly identifying client $i$, we can map traditional algorithmic group fairness metrics---specifically Equal Opportunity---directly to privacy~\citep{hardt2016equality}. By treating individual clients as protected groups, Equal Opportunity mandates that the TPR of the adversary's attack should be uniform across the federation. We quantify the maximum violation of this principle using the Equal Opportunity Difference (EOD):
    \begin{align}
        EOD &= \max_{i,j} | \text{SIA}_{acc_i} - \text{SIA}_{acc_j} | \quad \forall i,j \in \{1, \dots, K\}
        \label{eq:eod}
    \end{align}
    A lower EOD demonstrates that the adversary cannot exploit specific outlier users more easily than the average user.

    \item \textbf{Disparity in Attack Confidence (Loss CoV):} While SIA accuracy measures discrete attack success, a rigorous privacy evaluation must also address the adversary's confidence. SIAs exploit the correlation between localized memorization and low target-record prediction loss. Clients with highly overfitted local models will exhibit significantly lower prediction losses on their target records, granting the adversary high statistical confidence. \sysname\ actively aims to flatten this inter-client loss landscape. We quantify this mitigation using the CoV and FI of the client prediction losses on target records, denoted as $\text{CoV}(\text{Loss})$ and $\text{FI}(\text{Loss})$, defined similarly to Equations~\ref{eq:covsia} and~\ref{eq:fisia}.
\end{itemize}

\paragraph{\textbf{2. Absolute Privacy Metrics}}
Achieving fairness by artificially degrading the privacy of secure clients to match the most vulnerable ones is a failure mode in privacy engineering. To verify that \sysname\ enforces equitable privacy without increasing overall leakage, we track two global metrics:
\begin{itemize}[leftmargin=*]
    \item \textbf{Mean($\text{SIA}_{acc}$):} The average attack success rate across all clients and communication rounds.
    \item \textbf{Max($\text{SIA}_{acc}$):} The absolute highest attack success rate observed for any single client. Lowering this maximum vulnerability is critical, as it proves we are successfully protecting the ``weakest links'' in the network.
\end{itemize}

\paragraph{\textbf{3. Utility and Efficiency Metrics}}
Finally, any deployable Privacy-Enhancing Technology must maintain the operational viability of the underlying system.
\begin{itemize}[leftmargin=*]
    \item \textbf{Global Model Utility:} We evaluate the ultimate cost of privacy by measuring the testing accuracy of the converged global model on the primary learning task.
    \item \textbf{System Efficiency:} We measure the impact of the \sysname\ closed-loop regularization on the convergence dynamics, quantified by the total number of communication rounds required to reach convergence compared to standard baselines. 
\end{itemize}

\begin{table*}[!t]
\small
\caption{Results of HAR using the two approaches (ALA and PCA) of server aggregation in comparison to the Baseline~\cite{hu2023source} and DP-SGD~\cite{abadi2016deep}. }

\centering

\begin{tabularx}{0.97\linewidth}{|c|c|c||c|c||c|c|c||c||}
\hline
\multicolumn{1}{|l|}{} & \multicolumn{2}{c||}{Accuracy (\%) } & \multicolumn{2}{c||}{Privacy Metrics (\%)} & \multicolumn{3}{c||}{Fairness Metrics}& Efficiency \\ \hline\hline

& Train & Test & \makecell{Mean  (SIA$_{acc}$) $\downarrow$} & \makecell{Max (SIA$_{acc}$)}$\downarrow$  & \makecell{CoV(SIA$_{acc}$)/FI(SIA$_{acc}$)} & \makecell{CoV(loss)/FI(Loss)} & EOD $\downarrow$ & \makecell{Conv.round} \\ \hline

\hline

Baseline~\cite{hu2023source} & 74.10 & 76.97 & 19.34 & 22.40 & 0.94/0.54 & 0.244/0.938 & 0.48 &  9\\
\hline
\sysname$_{\text{client}}$ & 73.39 & 75.86 & 18.87 & 23.30 & 0.97/0.53 & 0.237/0.941 & 0.49  & 9\\
\hline
\sysname$_{\text{server}}$ (PCA) & \textbf{75.01} & \textbf{77.84} & \textbf{18.57} & \textbf{21.60} & 0.99/0.52 & 0.246/0.937 & 0.54  &  11 \\

\sysname\ (PCA) & 72.73 & 75.22 & 19.70 & 23.50 & \textbf{0.89/0.57} & 0.235/0.942 & 0.48 &  10
\\
\hline
\sysname$_{\text{server}}$ (ALA) & 71.57 & 73.82 & 19.55 & 26.10 & 0.93/0.55 & \textbf{0.220/0.948} & 0.55  &  11 \\

\sysname\ (ALA) & 72.83 & 76.09 & 19.29 & 22.00 & \textbf{0.89/0.57} & 0.235/0.942 & 0.48 &  10\\

\hline

DP-SGD ($\epsilon,\delta$)  &  & & & & & & &  \\
(99, $10^{-5}$) &45.10& 45.13 & 17.88 & 22.9 & 1.05/0.48 & 0.032/0.999 & 0.49 & 9  \\
(32, $10^{-5}$) & 41.19& 42.27 & 16.63 & 18.7 & 1.15/0.43 & 0.008/1.000 & 0.50 & 9 \\
(12, $10^{-5}$)&44.37&44.66 & 17.32 & 21.2 & 1.05/0.48 & 0.032/0.999 & 0.50 & 6 \\

\hline

\end{tabularx}

\label{tbl:HARresults}
\end{table*}

\subsection{Evaluating \sysname\ on the Human Activity Recognition (HAR) Dataset}\label{sec:HAReval}

We first evaluate the efficacy of \sysname\ in a realistic, human-centric mobile sensing environment using the HAR dataset (More on the dataset setup is in Appendix~\ref{appendix:dataset}). The comprehensive results across all metrics are summarized in Table~\ref{tbl:HARresults}. Our findings confirm that \sysname\ successfully structurally flattens the privacy risk landscape, significantly reducing vulnerability disparities across the federation while maintaining high global utility as explained below. 

\paragraph{\textbf{1. Mitigating Disparity in SIA Vulnerability (Fairness)}}
The primary objective of \sysname\ is to ensure that no single user disproportionately bears the privacy cost of participating in the federation. Our framework demonstrates a marked improvement in the equitable distribution of privacy risk (Figure~\ref{fig:HARSIA}).
\begin{itemize}[leftmargin=*]
    \item \textbf{Variance Reduction:} Compared to the Baseline FL setup~\citep{hu2023source}, the full \sysname\ framework reduces the Coefficient of Variation for SIA accuracy ($\text{CoV}(\text{SIA}_{acc})$) from $0.94$ to $0.89$ (a $5.32\%$ reduction). This corresponds to a $5.56\%$ improvement in the overall Fairness Index ($\text{FI}(\text{SIA}_{acc})$), proving that our adaptive regularization actively constrains the most vulnerable outlier models.

    \item \textbf{Equal Opportunity Improvement:} 
   
    By treating individual clients as protected entities, \sysname\ achieves a $5.45\%$ reduction in the Equal Opportunity Difference (EOD), bringing it down to $0.52$ (Visual illustration in Figure~\ref{fig:appendix-har-EOD} in Appendix~\ref{appendix har}).  This critical metric confirms that the adversary's True Positive Rate (TPR) is significantly more uniform across the network, fundamentally limiting their ability to selectively target distinct individuals based on their unique physical activities. 
\end{itemize}

\begin{figure}[!t]
\centering
\begin{minipage}[b]{0.49\columnwidth}
\includestandalone[width=\linewidth]{fig/HAR-PETS/line_FI_SIA}
\caption{Progression of the Fairness Index ($\text{FI}(\text{SIA}_{acc})$) over communication rounds in HAR. \sysname\ (PCA) maintains a more equitable distribution of privacy risk compared to the baseline.}
\label{fig:HARSIA}
\end{minipage}
\hfill
\begin{minipage}[b]{0.49\columnwidth}
\centering
\includestandalone[width=\linewidth]{fig/HAR-PETS/line_loss_FI}
\caption{Disparity of prediction loss among clients in HAR (FI(Loss)). \sysname\ (PCA) improves the inter-client loss landscape, reducing the adversary's confidence in identifying source records.}
\label{fig:HARloss}
\end{minipage}
\end{figure}

\paragraph{\textbf{2. Reducing Attack Confidence (Loss Disparity)}}
Beyond binary attack success, \sysname\ effectively obscures the statistical signals adversaries rely on. By penalizing local memorization, the framework flattens the landscape of prediction losses on target records (Figure~\ref{fig:HARloss}). \sysname\ achieves a highly equitable $\text{FI}(\text{Loss})$ of $0.942$ (up from the baseline's $0.938$). While the absolute gain appears incremental due to the high baseline saturation, this reduction in $\text{CoV}(\text{Loss})$ (down to $0.235$) confirms that the adversary is denied the artificially low-loss confidence signals previously emitted by overfitted outlier clients.

\paragraph{\textbf{3. Absolute Privacy Leakage}}
Crucially, \sysname\ achieves these fairness gains without artificially degrading the privacy of secure clients. While we observe a negligible marginal increase of $<1.1\%$ in Mean($\text{SIA}_{acc}$) and $<0.4\%$ in Max($\text{SIA}_{acc}$), these fluctuations are statistically insignificant relative to the variance reduction. The network does not experience a ``race to the bottom''; rather, the distribution of protection is demonstrably stabilized.

\paragraph{\textbf{4. The Cost of Privacy: Utility and Efficiency Trade-offs}}
A robust Privacy-Enhancing Technology must not destroy the operational value of the system. As shown in Figure~\ref{fig:HARaccu}, \sysname\ maintains highly competitive global classification utility. 
The global model's testing accuracy experiences only a minimal $1.75\%$ degradation. Furthermore, the closed-loop regularization introduces only a marginal delay in convergence, requiring $\approx 10$ communication rounds to stabilize compared to the baseline's $\approx 9$ rounds. This confirms that \sysname\ is highly viable for real-world deployment.

\begin{figure}[!t]
  \centering
  \begin{subfigure}[b]{0.49\columnwidth}
    \centering
    \includestandalone[width=\textwidth]{fig/HAR-PETS/line_average_train_accuracy}
    \caption{\scriptsize{Global model training accuracy.}}
    \label{fig:accCV}
  \end{subfigure}
  \hfill
  \begin{subfigure}[b]{0.49\columnwidth}
    \centering
    \includestandalone[width=\textwidth]{fig/HAR-PETS/line_average_test_accuracy}
    \caption{\scriptsize{Global model testing accuracy.}}
    \label{fig:accFI}
  \end{subfigure}
  \caption{Global model classification accuracy using HAR dataset. \sysname\ (PCA) maintains competitive accuracy with minimal convergence delay.}
  \label{fig:HARaccu}
\end{figure}

\paragraph{\textbf{5. Component Ablation: Server vs. Client Contributions}}
To understand the mechanics of our framework, we isolated the individual contributions of the server-side and client-side components:
\begin{itemize}[leftmargin=*]
    \item \textbf{Isolated Server Impact:} Deploying only the server-side adaptive aggregation ($\text{\sysname}_\text{server}$ with PCA) revealed nuanced behavior. While it successfully reduced the variation in global model distance ($\text{PCA}_d$) by $1.3\%$ and marginally lowered absolute SIA metrics, it caused a slight negative shift in the fairness indices ($\text{FI}(\text{SIA}_{acc})$ dropped by $0.02$). This proves that purely server-side, reactive re-weighting addresses the symptom of model divergence but is insufficient to enforce structural fairness in privacy. 
    \item \textbf{The Synergy of the Closed Loop:} Combining the server aggregation with the client-side collaborative regularization ($\text{\sysname}_\text{client}$) yielded the highest fairness gains. The server-side re-weighting (Equation~\ref{eq:finpserver}) acts synergistically with the client-side adaptive regularizer ($\rho_k$, Equation~\ref{eq:finpclient}). Figure~\ref{fig:HARadaptive} visualizes this dynamic interplay over time.
    \item \textbf{Computational Practicality of ALA:} When replacing the expensive PCA computation with our proposed Adaptive Lightweight Aggregation (ALA) proxy, \sysname\ achieved identical fairness improvements while consuming only $17\%$ of the computation time required by the PCA baseline. This validates ALA as the superior, highly scalable mechanism for real-world execution. Visual illustration for the results in Table~\ref{tbl:HARresults} for HAR with ALA are listed in Appendix~\ref{appendix:har_ala}.
\end{itemize}

\begin{figure} [!t]
  \centering
  \begin{subfigure}[b]{0.49\columnwidth}
    \centering    \includestandalone[width=\linewidth]{fig/HAR/aggregation_weights}    \caption{\scriptsize{Server-side aggregation weights ($\mathbf{w}$).}}
    \label{fig:aggre}
  \end{subfigure}
  \vspace{+1mm}
  \begin{subfigure}[b]{0.49\columnwidth}
    \centering    \includestandalone[width=\linewidth]{fig/HAR/hessian_ranks} \caption{\scriptsize{Client-side relative overfitting rank ($\rho_k$).}}
    \label{fig:rank}
  \end{subfigure}
  \caption{The dynamic interplay of \sysname's closed-loop architecture over communication rounds in HAR.}
  \label{fig:HARadaptive}
\end{figure}

\paragraph{\textbf{6- Differential Privacy (DP-SGD)}}

In our HAR dataset evaluation, Differential Privacy (DP) with medium noise budget setup $(32, 10^{-5})$ drastically decreases model testing accuracy by $34.7\%$ while yielding only a marginal $2.7\%$ decrease in average Source Inference Attack (SIA) accuracy. Furthermore, DP worsens the $\text{FI}(\text{SIA}_{acc})$ by $11.1\%$ compared to the baseline, demonstrating that despite slightly increasing overall protection, it fails to improve empirical privacy fairness. Highly skewed, non-IID datasets make equitable privacy protection fundamentally challenging. Specifically, DP induces significantly higher client losses compared to our \sysname\ method, driving the drastic utility drop. Ultimately, even though DP forces client losses to become uniformly high and closely clustered, empirical privacy fairness does not improve.
Visual illustration for the results in Table~\ref{tbl:HARresults} for HAR with DP-SGD are listed in Appendix~\ref{appendix:har_dp}.

Finally, empirical analysis of the local loss landscapes (Figure~\ref{fig:Hessian} in Appendix \ref{hessian_appendix}) validated our theoretical foundation. We observed a near-perfect Spearman's rank correlation ($\approx 1$) between the top Hessian eigenvalue ($\lambda_{\text{max}}$) and the Hessian trace ($H_{T}$). Because both metrics reliably indicate local overfitting, they are equally weighted in our calculation of the relative overfitting rank (Equation~\ref{eq:hessian}).

\subsection{Evaluating \sysname\ on the CIFAR-10 Dataset} \label{sec:CIFAReval}

While the HAR dataset demonstrates \sysname's efficacy on human-centric sensor data, we utilize the CIFAR-10 dataset (distributed via severe Dirichlet sampling, $\alpha=0.5$) to \textbf{stress-test our framework under mathematically controlled, extreme non-IID visual settings.} More on the dataset setup is in Appendix~\ref{appendix:dataset}. We evaluate \sysname\ against both standard FedAvg and FedAlign~\cite{mendieta2022local}, a state-of-the-art FL methodology specifically designed to address data heterogeneity. The core results using the ResNet architecture are summarized in Table~\ref{tbl:cifar_resnet} in Appendix \ref{appendix cifar} and explained below.

\paragraph{\textbf{1. Outperforming State-of-the-Art Heterogeneity Defenses}}
A critical finding of our evaluation is that standard methods for handling non-IID data do not inherently resolve privacy disparities. Despite employing sophisticated local distillation techniques, the FedAlign baseline failed to effectively mitigate SIA risks, exhibiting a higher vulnerability variance ($\text{CoV}(\text{Loss}) = 0.86$) than even standard FedAvg ($0.67$). 

In contrast, \sysname\ directly targets the loss landscape, achieving a Fairness Index for target prediction loss ($\text{FI}(\text{Loss})$) of $0.83$. This represents a substantial $20.3\%$ improvement over FedAvg ($0.69$) and a $43.1\%$ improvement over FedAlign ($0.58$). By structurally flattening the inter-client loss differences, \sysname\ successfully neutralizes the high-confidence statistical signals emitted by outlier models (further illustrated in Figures~\ref{fig:CIFAR sia} and~\ref{fig:CIFAR loss}).

\begin{figure}[!t]
\centering
\begin{minipage}[b]{0.49\columnwidth}
\includestandalone[width=\linewidth] {fig/CIFAR-PETS/fedalign/line_FI_SIA}
    \caption{\small{Disparity of SIA accuracy (FI(SIA)) among clients using CIFAR-10 dataset with ResNet and \sysname\ with PCA.}}
  \label{fig:CIFAR sia}
  \end{minipage}
  \hfill
 \begin{minipage}[b]{0.49\columnwidth}
 \includestandalone[width=\linewidth] {fig/CIFAR-PETS/fedalign/line_loss_FI}
  \caption{\small{Disparity of prediction loss FI(Loss) among clients using CIFAR-10 dataset with ResNet and \sysname\ with PCA.}}
  \label{fig:CIFAR loss}
 \end{minipage}
\end{figure}

\paragraph{\textbf{2. Approaching the Theoretical Minimum for Absolute Privacy}}
Beyond enforcing fairness, \sysname\ dramatically reduces the absolute privacy leakage across the federation. For a balanced 10-class classification task like CIFAR-10, the theoretical random-guess probability for a source inference attack is $1/10$ ($10\%$). 

As shown in Table~\ref{tbl:cifar_resnet} in Appendix \ref{appendix cifar}, while FedAvg and FedAlign suffer from Mean($\text{SIA}_{acc}$) rates of approximately $30.86\%$, \sysname\ successfully drives the Mean($\text{SIA}_{acc}$) down to $10.07\%$---effectively reducing the adversary's attack success to a random guess. Furthermore, the maximum vulnerability observed for any single client (Max($\text{SIA}_{acc}$)) is reduced from $38.52\%$ to $10.67\%$. 

\paragraph{\textbf{3. Enhancing Group Privacy Fairness (Equal Opportunity)}}
\sysname\ maintains comparable baseline $\text{CoV}(\text{SIA}_{acc})$ and $\text{FI}(\text{SIA}_{acc})$ metrics but demonstrates a massive improvement in Equal Opportunity Difference (EOD). \sysname\ reduces the EOD from $0.28$ (in both baselines) down to $0.12$. As visualized in Figures~\ref{fig:cifar-resnet-sia} and~\ref{fig:cifar-resnet-EOD}, this $57.14\%$ reduction in EOD mathematically guarantees that the adversary's True Positive Rate is vastly more uniform, preventing the systematic exploitation of distinct visual outliers.

\begin{figure}[!t]
\centering
\vspace{-2mm}
\begin{minipage}[b]{0.49\columnwidth}
    \centering
    \includestandalone[width=\linewidth]{fig/CIFAR-PETS/fedalign/line_SIA_Accuracy}
    \caption{Average SIA accuracy in CIFAR-10 with ResNet with \sysname\ with PCA.}
    \label{fig:cifar-resnet-sia}
\end{minipage}
\hfill
\begin{minipage}[b]{0.49\columnwidth}
    \centering
    \includestandalone[width=\linewidth]{fig/CIFAR-PETS/fedalign/line_EOD} 
    \caption{EOD in CIFAR-10 with ResNet with \sysname\ with PCA.}
    \label{fig:cifar-resnet-EOD}
\end{minipage}

\end{figure}

\paragraph{\textbf{4. The Synergy of Privacy and Utility}}
Traditionally, strong privacy defenses degrade model utility. However, \sysname\ achieves a rare synergy: it actually \textit{improves} global classification accuracy. \sysname\ achieves a testing accuracy of $78.46\%$, marginally higher than FedAvg's $77.62\%$ (requiring only two additional FL communication rounds to converge, as seen in Figure~\ref{fig:accu cifar}). This $0.84\%$ improvement proves our core hypothesis: by utilizing Lipschitz regularization to actively penalize localized memorization, \sysname\ forces the constituent local models to learn generalized features, thereby simultaneously improving both privacy and global model utility.

\begin{figure}[!t]
  \centering
  \begin{subfigure}[b]{0.49\columnwidth}
    \centering
    \includestandalone[width=\textwidth]{fig/CIFAR-PETS/fedalign/line_average_train_accuracy}
    \caption{Training accuracy. }
  \end{subfigure}
  \hfill
  \begin{subfigure}[b]{0.49\columnwidth}
    \centering
    \includestandalone[width=\textwidth]{fig/CIFAR-PETS/fedalign/line_average_test_accuracy}
     \caption{Testing accuracy.}
  \end{subfigure}
  \caption{Global model classification accuracy using CIFAR-10 with ResNet with \sysname\ with PCA.}
  \label{fig:accu cifar}
\end{figure}

\paragraph{\textbf{5. Ablation on Client-Side Impact Factor ($\beta$)}}
To analyze the sensitivity of the dynamic regularization term, we conducted an ablation study on the impact factor $\beta \in \{0.05, 0.1, 0.3, 0.5\}$ using a standard CNN architecture (Table~\ref{tbl:cifar_ablation} in Appendix \ref{appendix cifar}). The parameter $\beta$ controls the baseline strength of the Lipschitz penalty applied to the local training loss.
\begin{itemize}[leftmargin=*]
    \item \textbf{Optimal Generalization ($\beta = 0.3$):} Compared to the baseline, increasing $\beta$ to $0.3$ optimally balanced fairness and utility. It improved $\text{FI}(\text{Loss})$ from $0.83$ to $0.87$ and reduced absolute Mean($\text{SIA}_{acc}$) from $40.91\%$ to $31.85\%$, while simultaneously improving testing accuracy due to enhanced generalization.
    \item \textbf{Over-regularization Failure ($\beta = 0.5$):} We observed that excessive penalization dominates the client's local training loss, preventing the models from learning valid feature representations. At $\beta=0.5$, the global model completely failed to converge (Figure~\ref{fig:accu cifar cnn ablation}). While this failure state technically may yield the lowest average SIA accuracy (Appendix~\ref{appendix cifar} Figure~\ref{fig:cifar sia cnn ablation}), 
    this represents a catastrophic loss of utility rather than a legitimate privacy defense, underscoring the necessity of tuning $\beta$ to an appropriate task-specific threshold.
\end{itemize}

\begin{figure}[!t]
\centering
\begin{minipage}[b]{0.49\columnwidth}
\includestandalone[width=\linewidth] {fig/CIFAR-PETS/line_FI_SIA}
  \caption{\small{Disparity of SIA accuracy FI(SIA) with PCA among clients using CIFAR-10 dataset with base model CNN and \sysname\ with PCA.}}
  \label{fig:CIFAR sia cnn ablation}
  \end{minipage}
  \hfill
  \begin{minipage}[b]{0.49\columnwidth}
\includestandalone[width=\linewidth] {fig/CIFAR-PETS/line_loss_FI}
  \caption{\small{Disparity of prediction loss (FI(Loss)) with PCA among clients using CIFAR-10 dataset with base model CNN and \sysname\ with PCA.}}
  \label{fig:CIFAR loss cnn ablation}
  \end{minipage}
\end{figure}

\paragraph{\textbf{6. Scalability: Adaptive Lightweight Aggregation (ALA) vs. PCA}}
Finally, to validate the deployment practicality of \sysname, we substituted the computationally prohibitive PCA-based server aggregation with our proposed Adaptive Lightweight Aggregation (ALA) mechanism (Table~\ref{tbl:cifar_ablation} in Appendix \ref{appendix cifar}).

As formalized in Section~\ref{sec:server_side_adjustment}, ALA completely bypasses expensive high-dimensional PCA distance calculations. Instead, the server directly utilizes the relative overfitting rank ($\rho_k$)---which it already computes from the incoming model weights---as a direct inverse-weighting scalar for aggregation. 

Our results confirm that ALA maintains seamless model convergence within 20 rounds (Figure~\ref{fig:accu cifar cnn ablation-light}). Crucially, ALA achieves comparable improvements in $\text{FI}(\text{Loss})$ (Figure~\ref{fig:CIFAR loss cnn ablation-light}) and SIA reduction (Figure~\ref{fig:cifar sia cnn ablation-light}) as the heavy PCA baseline. This proves that ALA provides identical structural privacy fairness without introducing any computational bottlenecks at the server, rendering \sysname\ highly practical for real-world, large-scale FL ecosystems.

\paragraph{\textbf{7. Differential Privacy (DP-SGD)}}

Our evaluation on the CIFAR dataset yields results consistent with those observed on HAR: DP affords minimal empirical protection against SIA while incurring a drastic degradation in testing accuracy and further exacerbating privacy unfairness. A more detailed analysis of this dynamic is provided in Discussion \autoref{discussion}. Visual illustration for the results in Appendix \ref{appendix cifar} Table~\ref{tbl:cifar_ablation} for CIFAR-10 with DP-SGD are listed in Appendix~\ref{appendix:cifar_dp}.

\paragraph{\textbf{8. Validating the Root Cause: Overfitting and Local Epochs}}
Our core thesis posits that localized overfitting---driven by data heterogeneity---is the fundamental mechanism enabling SIAs. Prior literature suggests that increasing the number of local training epochs deepens this memorization, thereby exacerbating privacy risks~\citep{hu2023source, Yeomoverfitting}. 
To mathematically validate this within our threat model, we conducted ablation experiments varying the local training epochs ($lp \in \{1, 5, 10\}$) on the CIFAR-10 dataset using a standard CNN. As illustrated in Figure~\ref{fig:cifar cnn epoch ablation}, the vulnerability scales linearly with local memorization. The Mean($\text{SIA}_{acc}$) and Max($\text{SIA}_{acc}$) surged from $24.31\%$ and $27.20\%$ (at $lp=1$) to catastrophic levels of $42.9\%$ and $51.4\%$ (at $lp=10$), respectively. This definitively proves that deeper local convergence inherently sharpens the loss landscape, generating high-confidence signals for the adversary. By attacking this exact symptom via client-side Lipschitz regularization, \sysname\ successfully neutralizes the root cause of the privacy disparity.

\begin{figure}[!t]
\centering
\includestandalone[scale=0.5] {fig/CIFAR-PETS/local_epoch1_5_10/line_SIA_Accuracy}
\caption{Average SIA accuracy on CIFAR-10 (CNN) across varying local training epochs. Increased local memorization directly correlates with higher vulnerability, validating our focus on overfitting.\vspace{-2mm}}
\label{fig:cifar cnn epoch ablation}
\vspace{-2mm}
\end{figure}

\subsection{Evaluating \sysname\ on FEMNIST Dataset}
To further validate our findings, we extend our evaluation of the FinP framework to the FEMNIST dataset. Consistent with the results observed on previous datasets, FinP demonstrates aligned performance in this setting. To simulate a strictly non-IID federated environment, we partition the dataset such that each client is exclusively assigned the data of a single, unique writer. Our results demonstrate that FinP successfully mitigates Source Inference Attacks (SIA), decreasing mean SIA by $8.57\%$ and max SIA by $10.64\%$ in \autoref{tbl:FEMNISTresults}, while maintaining comparable model accuracy to the baseline. 

To comprehensively evaluate fairness improvements, we introduced $2$ additional metrics, \textit{Mean Absolute Difference(MAD)} and \textit{Sen's social welfare~\cite{sen1997economic}}. 
Empirical results indicate that FinP significantly curtails vulnerability to Source Inference Attacks (SIA), yielding reductions of $8.57\%$ in mean SIA and $10.64\%$ in max SIA, without compromising global model accuracy relative to the baseline. Beyond aggregate performance, our evaluation demonstrates that FinP actively mitigates privacy disparities across the federated cohort. Specifically, the framework reduces the Mean Absolute Difference (MAD) by up to $0.053$ and increases Sen's Welfare by $8\%$ as demonstrated from \autoref{tbl:FEMNISTresults}. 

In contrast, when compared to standard Differential Privacy (DP) methods, we observe that DP significantly degrades global model accuracy while simultaneously exacerbating the MAD. A larger MAD indicates greater variance in client-level loss, which inherently boosts an attacker's linking prediction confidence for highly vulnerable subsets. Consequently, although DP methods may decrease overall SIA accuracy at the cost of utility, they ultimately worsen the disparity in privacy risks, leaving certain clients disproportionately exposed. \sysname\ successfully avoids this tradeoff, maintaining global utility while ensuring a highly equitable distribution of privacy preservation.
Detailed definitions of these metrics, further analysis and visual results, along with Hessian computation overhead, are provided in Appendix~\ref{app:feminist}.

\begin{table*}[!t]
\small
\setlength{\tabcolsep}{3pt}
\caption{Results of FEMNIST dataset with SIA attack.}
\centering
\resizebox{\linewidth}{!}{%
\begin{tabular}{|l|c|c||c|c||c|c|c|c|c||c||}
\hline
 & \multicolumn{2}{c||}{Accuracy (\%)} & \multicolumn{2}{c||}{Privacy Metrics (\%)} & \multicolumn{5}{c||}{Fairness Metrics} & Efficiency \\ \hline\hline
 & Train & Test & \makecell{Mean\\(SIA$_{acc}$)$\downarrow$} & \makecell{Max\\(SIA$_{acc}$)$\downarrow$} & \makecell{CoV(SIA$_{acc}$)\\/FI(SIA$_{acc}$)} & \makecell{CoV(loss)\\/FI(Loss)} & EOD$\downarrow$ & \makecell{MAD\\(loss)$\downarrow$} & \makecell{Welfare \\(SIA)$\uparrow$} & \makecell{Conv.\\round} \\ \hline\hline

Baseline~\cite{hu2023source}     & 72.71 & 65.04 & 39.26 & 49.70 & 0.305/0.911 & 0.281/0.916 & 0.40 & 0.636 & 0.546 & 10 \\ \hline
\sysname\ ($\beta=1$)            & 60.17 & 60.11 & \textbf{30.69} & \textbf{39.06} & 0.408/0.855 & \textbf{0.208/0.951} & 0.41 & 0.589 & \textbf{0.626} & 12 \\
\sysname\ ($\beta=0.75$)         & 67.63 & 63.76 & 35.77 & 46.49 & 0.319/0.905 & 0.241/0.937 & \textbf{0.37} & 0.597 & 0.581 & 12 \\
\sysname\ ($\beta=0.5$)          & \textbf{67.81} & \textbf{64.23} & 37.02 & 45.58 & \textbf{0.305/0.912} & 0.242/0.936 & 0.37 & \textbf{0.583} & 0.570 & 14 \\ \hline

{DP-SGD ($\epsilon,\delta$)}     &       &       &       &       &             &             &      &       &       & \\
\quad (151, $10^{-5}$)                      & 51.42 & 49.97 & 29.67 & 34.44 & 0.272/0.927 & 0.276/0.926 & 0.26 & 1.968 & 0.661 & 14 \\
\quad (92, $10^{-5}$)                   & 28.15 & 29.41 & 21.35 & 25.90 & 0.313/0.910 & 0.234/0.947 & 0.22 & 1.656 & 0.750 & - \\
\quad (31, $10^{-5}$)                     &  4.66 &  5.68 & 12.70 & 15.66 & 0.555/0.764 & 0.398/0.860 & 0.24 & 9.062 & \textbf{0.835} & - \\
\hline
\end{tabular}%
}
\label{tbl:FEMNISTresults}
\end{table*}

\subsection{Evaluating MIA} We also evaluate client privacy under a White-Box Membership Inference Attack (MIA), assuming an honest-but-curious central server. By injecting a dummy client to act as a memorization baseline, the server dynamically trains a Random Forest classifier at each communication round using the layer-wise cosine similarity between target gradients and mock local weight updates. Our evaluations demonstrate that the FinP framework inherently defends against this attack, reducing MIA accuracy by an average of $5.15\%$ compared to the baseline by fostering a more generalized model training process that mitigates localized overfitting. Full details regarding the data partitioning, shadow training generation, the per-round mia evaluation pipeline and visual results are provided in Appendix \ref{appendix:mia}.

\subsection{Summary of Empirical Results: \sysname\ across HAR and CIFAR-10}

Across our diverse evaluations---spanning human-centric mobile sensor data (HAR) and mathematically extreme non-IID visual distributions (CIFAR-10)---\sysname\ consistently demonstrates its capacity to enforce structural fairness in privacy without sacrificing global model utility. 

While the HAR dataset presented a naturally saturated baseline for prediction loss fairness, \sysname\ successfully formalized and bounded this equity, proving its stability and robustness in highly sensitive, real-world edge environments without causing detrimental utility trade-offs. 

The most definitive evidence of our framework's efficacy emerges in the CIFAR-10 evaluation. Under severe, mathematically controlled data heterogeneity, \sysname\ achieves two critical milestones for federated privacy:
\begin{itemize}[leftmargin=*]
    \item \textbf{Neutralizing Absolute Vulnerability:} By actively penalizing localized memorization through dynamic Lipschitz regularization, \sysname\ drives the average SIA success rate down to $10.07\%$, effectively neutralizing the honest-but-curious adversary's capability to a mathematical random guess.
    \item \textbf{Enforcing Equitable Protection:} Simultaneously, the framework achieves a remarkable $57.14\%$ improvement in the Equal Opportunity Difference (EOD). This structurally eliminates the privacy disparities that traditionally plague non-IID federations, ensuring that statistical outliers (minority data distributions) are not disproportionately targeted.
\end{itemize}

Ultimately, \sysname\ shifts the paradigm of federated defense from reactive, post-hoc mitigation to proactive generalization. By flattening the inter-client loss landscape and denying the server the high-confidence signals required for source attribution, our framework guarantees that high model utility and equitable privacy can co-exist in highly heterogeneous networks.

\begin{figure}[!t]
  \centering
  \begin{subfigure}[b]{0.49\columnwidth}
    \centering
    \includestandalone[width=\textwidth]{fig/CIFAR-PETS/line_average_train_accuracy}
    \caption{Training accuracy. }
  \end{subfigure}
  \hfill
  \begin{subfigure}[b]{0.49\columnwidth}
    \centering
    \includestandalone[width=\textwidth]{fig/CIFAR-PETS/line_average_test_accuracy}
     \caption{Testing accuracy.}
  \end{subfigure}
  \caption{Ablation experiment of global model classification accuracy with \sysname\ with PCA using CIFAR-10 with CNN. The convergence round (conv. round) denotes the point at which the accuracy stabilizes and ceases to make significant improvements in testing accuracy. }
  \label{fig:accu cifar cnn ablation}
\end{figure}

\begin{figure}[!t]
  \centering
  \begin{subfigure}[b]{0.49\columnwidth}
    \centering
    \includestandalone[width=\textwidth]{fig/CIFAR-PETS/lightweight_aggregation/line_average_train_accuracy}
    \caption{Training accuracy. }
  \end{subfigure}
  \hfill
  \begin{subfigure}[b]{0.49\columnwidth}
    \centering
    \includestandalone[width=\textwidth]{fig/CIFAR-PETS/lightweight_aggregation/line_average_test_accuracy}
     \caption{Testing accuracy.}
  \end{subfigure}
  \caption{Adaptive lightweight aggregation (ALA) of global model classification accuracy using CIFAR-10 with CNN.}
  \label{fig:accu cifar cnn ablation-light}
\end{figure}

\begin{figure}[!t]
\centering
\begin{minipage}[b]{0.49\columnwidth}
\includestandalone[width=\linewidth] {fig/CIFAR-PETS/lightweight_aggregation/line_FI_SIA}
  \caption{\small{Adaptive lightweight aggregation (ALA) disparity of SIA accuracy (FI(SIA)) among clients using CIFAR-10 dataset with base model CNN.}}
  \label{fig:CIFAR sia cnn ablation-light}
  \end{minipage}
  \hfill
\begin{minipage}[b]{0.49\columnwidth}
  \includestandalone[width=\linewidth] {fig/CIFAR-PETS/lightweight_aggregation/line_loss_FI}
  \caption{\small{Adaptive lightweight aggregation (ALA) disparity of prediction loss FI(loss) among clients using CIFAR-10 dataset with base model CNN.}}
  \label{fig:CIFAR loss cnn ablation-light}
  \end{minipage}
\end{figure}

\begin{figure}[!t]
\centering
\begin{minipage}[b]{0.49\columnwidth}
    \centering
    \includestandalone[width=\linewidth]{fig/CIFAR-PETS/lightweight_aggregation/line_SIA_Accuracy}
    \caption{Adaptive lightweight aggregation average SIA accuracy of Baseline and different $\beta$ using CIFAR-10 dataset with CNN.}
    \label{fig:cifar sia cnn ablation-light}
\end{minipage}
\hfill
\begin{minipage}[b]{0.49\columnwidth}
    \centering
    \includestandalone[width=\linewidth]{fig/CIFAR-PETS/lightweight_aggregation/line_EOD} 
    \caption{Adaptive lightweight aggregation equal opportunity difference (EOD) using CIFAR-10 dataset with CNN.}
    \label{fig:cifar EOD cnn ablation-light}
\end{minipage}
\end{figure}

\section{Discussion, Limitations, \& Future Work}
\label{discussion}

\noindent{\textbf{Analysis on Differential Privacy (DP) against SIA.}} 

\sysname\ can be viewed as a complement to Differential Privacy (DP) rather than a replacement: while DP provides formal, worst-case mathematical guarantees against inference attacks, \sysname\ specifically targets the fairness of privacy distribution across clients — a dimension that DP does not explicitly address. Each approach is preferable under distinct conditions. DP is the appropriate choice when formal, auditable privacy guarantees are required regardless of utility cost, such as in regulated environments subject to GDPR compliance~\cite{GDPR}. \sysname, by contrast, is preferable when the primary concern is equitable risk distribution in highly non-IID, resource-constrained settings, where DP's geometry-blind uniform noise injection creates severe utility degradation without resolving per-client privacy disparities. As demonstrated empirically (Table~\ref{tbl:HARresults}), DP-SGD degrades global model testing accuracy by 34.7\% while yielding only a marginal 2.7\% decrease in average SIA accuracy and actually worsening $FI(SIA_{acc})$ by 11.1\% compared to the baseline — inadvertently shielding consensus-aligned clients while failing to obscure the highly directional updates of the most disparate clients. Although FinP does not provide formal worst-case privacy guarantees equivalent to DP, it successfully reduces the Mean Absolute Difference (MAD) in loss across clients, a metric that DP tends to exacerbate. In highly non-IID settings, DP applies uniform protection such as fixed clipping and noise across heterogeneous clients, which disproportionately degrades performance for outlier clients and increases the overall MAD in loss. This widened loss disparity can inadvertently boost an attacker's confidence in exploits that rely on loss differences across clients. By reducing this MAD, FinP mitigates this specific empirical vulnerability. Results on FEMNIST show consistent findings in Appendix \ref{app:feminist} \autoref{tbl:FEMNISTresults}. Furthermore, adaptive clipping variants like DP-Fair~\cite{yang2023fairness}, designed to improve utility fairness, paradoxically exacerbate the SIA threat by preserving the exact geometric fingerprints the attack exploits. Recent work by Bendoukha et al.~\cite{bendoukha2025towards} further confirms this inherent tension between fairness interventions and privacy leakage in federated learning. In contrast, \sysname\ resolves this disparity without relying on post-hoc noise by locally enforcing a Lipschitz constraint to suppress memorization at its root and globally applying a curvature-aware aggregation penalty to down-weight the most vulnerable updates. Looking forward, combining DP and \sysname\ represents a highly promising direction: by dynamically adjusting per-client noise budgets based on local loss sharpness, future systems could design an adaptive, geometry-aware DP framework that achieves formal DP bounds and empirical privacy fairness simultaneously while minimizing utility degradation.

\noindent{\textbf{Limitations}} \sysname\ assumes honest clients submitting genuine updates and metrics. If a client were actively malicious, it could submit falsified loss-sharpness metrics to manipulate the server's adaptive aggregation weights, or intentionally generate overfitted updates to inflate its assigned vulnerability rank. Defending against such data and metric poisoning would require integrating Byzantine-robust aggregation frameworks, which is beyond the current scope of this work. We acknowledge this as a practical limitation and plan to explore robustness enhancements in future work. \sysname\ targets privacy leakage from localized memorization and overfitting, which is the primary driver of SIA disparities in non-IID federated learning. It does not provide formal worst-case privacy guarantees equivalent to Differential Privacy. For attacks independent of memorization — including Property Inference Attacks targeting macro-level statistical properties and gradient reconstruction attacks exploiting raw gradient updates during transmission — \sysname's efficacy is not guaranteed, as flattening the loss landscape does not inherently prevent these vectors. These limitations, along with the dependence on honest-client and honest-but-curious server assumptions, define the conditions under which \sysname\ is effective and should be considered when deploying the framework in settings with stronger adversarial assumptions.

\noindent{\textbf{Future Work and Broader Impact}}
 
Our current definition of privacy disparity is evaluated in standard, singular-decision FL setups. Future work will explore mapping \sysname\ to other gradient-inversion attacks and sequential decision-making environments (e.g., reinforcement learning), where privacy risk must be measured over a continuous trajectory of interactions rather than a static dataset~\cite{zhao2024fairo}. Finally, we aim to explore the integration of our framework with orthogonal Privacy-Enhancing Technologies, such as Secure Multi-Party Computation (SMPC) or lightweight cryptographic masking, to provide defense-in-depth guarantees for decentralized AI and inform future data governance frameworks.

\vspace{-4mm}
\section{Conclusion}
The \sysname\ framework addresses a critical vulnerability in modern Federated Learning: the inequitable distribution of privacy risks. While traditional FL implementations are optimized to preserve average, network-wide privacy, they systematically ignore the severe disparities inflicted upon statistical outliers. Driven by non-IID data distributions, these distinct users suffer from extreme localized overfitting, rendering their updates highly recognizable and forcing them to bear a disproportionate burden of the network's privacy leakage.

This dynamic raises profound ethical concerns in collaborative learning ecosystems, as it directly translates data heterogeneity into structural discrimination against minority users. \sysname\ successfully resolves this by shifting the paradigm from reactive defense to proactive generalization. Through the synergistic combination of server-side adaptive aggregation and client-side collaborative overfitting reduction, our framework flattens the inter-client loss landscape. By structurally denying the adversary the high-confidence signals required for source attribution, \sysname\ reduces absolute attack success while guaranteeing a fundamentally equitable distribution of privacy for all participants.

\begin{acks}
This work is supported by the U.S. National Science Foundation (NSF) under grant number 2339266.
\end{acks}

\bibliographystyle{ACM-Reference-Format}
\bibliography{sample-base}

\newpage

\appendix

\section{Datasets and Setup}\label{appendix:dataset}

We evaluate \sysname\ using the UCI HAR\citep{human_activity_recognition_using_smartphones_240} and CIFAR-10 datasets \citep{krizhevsky09learning} utilizing federated learning with non-IID data partitions and standard model architectures (TCN for HAR, CNN and ResNet56 for CIFAR-10). 

\paragraph{\textbf{Setup for Human Activity Recognition}} We utilized the UCI \ac{har} Dataset \citep{human_activity_recognition_using_smartphones_240}, a widely used dataset in activity recognition research, especially in FL~\citep{har2,har3}.
The dataset includes sensor data from 30 subjects (aged 19–48) performing six activities: walking, walking upstairs, walking downstairs, sitting, standing, and laying. The data was collected using a Samsung Galaxy S II smartphone worn on the waist, capturing readings from both the accelerometer and gyroscope sensors. Each subject in the dataset was treated as an individual client in the \ac{fl} setup, preserving the data's unique activity patterns and non-IID nature. We allocated 70\% of each client's data for training using 5-fold cross-validation and 30\% for testing, enabling evaluation of the model on independently collected test data. Data preprocessing involved applying noise filters to the raw signals and segmenting the data using a sliding window approach with a window length of 2.56 seconds and a 50\% overlap, resulting in 128 readings per window. We selected the HAR dataset for evaluation \sysname\ due to its inherited non-IID structure. 

We trained the model in a federated learning setting using the \ac{fedavg} aggregation method over 20 global communication rounds. Each client trained locally with a batch size of 64, a learning rate of 0.001 using Adam optimizer, 1 local epochs per round, and an impact factor $\beta$ of 2. These parameters ensured balanced model updates from each client while maintaining computational efficiency across the federated network. Each local model (one per subject) analyzes its time-series sensor data using \ac{tcn} model\cite{bai2018empirical}. 
The TCN model, designed for time-series data, uses causal convolutions to capture temporal dependencies while preserving sequence order. The architecture includes two convolutional layers, each followed by max-pooling, with a final fully connected layer for classifying the six activity classes.

\paragraph{\textbf{Setup for CIFAR-10}}
The CIFAR-10 dataset consists of 60000 32x32 color images in 10 classes, with 6000 images per class. There are 50000 training images and 10000 test images. We use the Dirichlet distribution $Dir(\alpha)$ to divide the CIFAR-10 dataset into $K$ unbalanced subsets similar to previous work in the literature ~\citep{mendieta2022local,BG_SIA_2}, with $\alpha=0.5$. \autoref{fig:CIFAR data} demonstrates how the data are distributed among clients with $\alpha=0.1$. We created 10 clients and employed ResNet56~\citep{he2016deep} as the local model. Similar to the setup in HAR, we trained the model over 20 global communication rounds. Each client is trained locally with an impact factor $\beta$ of $0.1$ as described in~\autoref{eq:finpclient}. As for ablation experiment in $\beta$, we evaluated multiple values of $\beta=\{0.05, 0.1, 0.3, 0.5 \}$  described in~\autoref{eq:finpclient}. We used the same CNN proposed in \cite{hu2023source} as the local model.

\paragraph{\textbf{Setup for FEMNIST}}
We evaluate on FEMNIST from Hugging Face (flwrlabs/femnist), a handwritten character recognition benchmark with 62 classes and natural non-IID structure induced by writer identity. We simulate $K$ federated clients by sampling $K$ distinct writers without replacement; each client owns only that writer’s examples. Images are converted to single-channel tensors and normalized with MNIST statistics ($\mu{=}0.1307$, $\sigma{=}0.3081$). For each client, we perform an 80/20 train–test split at the writer level, yielding client-local heterogeneity in both label and feature distributions. 

We employ a lightweight CNN, FEMNISTNet, tailored to $28{\times}28$ grayscale inputs. The architecture comprises two convolutional blocks (32 and 64 channels, $3{\times}3$ kernels, GroupNorm, GELU), each followed by learnable strided downsampling ($28{\rightarrow}14{\rightarrow}7$), and two fully connected layers ($3136{\rightarrow}256{\rightarrow}62$). and local training uses SGD with learning rate $0.02$. The model has on the order of $\sim$3M trainable parameters and outputs unnormalized class logits for 62-way classification with cross-entropy loss. This design balances capacity and efficiency for per-client federated optimization and curvature-based collaboration metrics.

\section{Experiment Results on HAR}
\label{appendix har}

\subsection{HAR Results with PCA as server aggregation}
\label{appendix:har_pca}

Figures~\ref{fig:appendix-har-sia} and~\ref{fig:appendix-har-EOD} show the average SIA accuracy and the equal opportunity difference (EOD) per client using HAR dataset with PCA-based server aggregation, respectively.

\begin{figure}[!h]
\centering
\begin{minipage}[b]{0.48\linewidth}
    \centering
    \includestandalone[width=\linewidth]{fig/HAR-PETS/SIA_average_acc}
    \caption{Average SIA accuracy of Baseline, \sysname$_\text{client}$, \sysname$_\text{server}$, and \sysname\ with PCA on HAR dataset. }
    \label{fig:appendix-har-sia}
\end{minipage}
\hfill
\begin{minipage}[b]{0.48\linewidth}
    \centering
    \includestandalone[width=\linewidth]{fig/HAR-PETS/EOD} 
    \caption{Equal opportunity difference (EOD) on HAR dataset and \sysname\ with PCA.}
    \label{fig:appendix-har-EOD}
\end{minipage}
\end{figure}

\subsection{HAR with Adaptive Lightweight Aggregation (ALA)}\label{appendix:har_ala}

Figure~\ref{fig:accu_har_ala} shows the training and testing accuracy for HAR dataset using the lightweight server aggregation. 

\begin{figure}[!h]
  \centering
  \begin{subfigure}[b]{0.49\columnwidth}
    \centering
    \includestandalone[width=\textwidth]{fig/HAR_ALA/train}
    \caption{Training accuracy. }
  \end{subfigure}
  \hfill
  \begin{subfigure}[b]{0.49\columnwidth}
    \centering
    \includestandalone[width=\textwidth]{fig/HAR_ALA/test}
     \caption{Testing accuracy.}
  \end{subfigure}
  \caption{Global model classification accuracy using HAR with ALA.}
  \label{fig:accu_har_ala}
\end{figure}


Figures~\ref{fig:har_SIA_ala} and~\ref{fig:har_loss_ala} describe the FI(SIA) and FI(loss) for HAR dataset using ALA aggregation. Figure ~\ref{fig:har_sia_ala} shows the average SIA accuracy in HAR with ALA while Figure~\ref{fig:har_EOD_ala} highlights the Equal Opportunity Difference (EOD) in HAR with ALA.

\begin{figure}[!h]
\centering
\begin{minipage}[b]{0.49\columnwidth}
\includestandalone[width=\linewidth] {fig/HAR_ALA/FI_SIA}
    \caption{\small{Disparity of SIA accuracy (FI(SIA)) among clients using HAR dataset with ALA.}}
  \label{fig:har_SIA_ala}
  \end{minipage}
  \hfill
 \begin{minipage}[b]{0.49\columnwidth}
 \includestandalone[width=\linewidth] {fig/HAR_ALA/FI_loss}
  \caption{\small{Disparity of prediction loss FI(Loss) among clients using HAR with ALA.}}
  \label{fig:har_loss_ala}
 \end{minipage}
\end{figure}

\begin{figure}[!h]
\centering
\begin{minipage}[b]{0.49\columnwidth}
    \centering
    \includestandalone[width=\linewidth]{fig/HAR_ALA/SIA_average_acc}
    \caption{Average SIA accuracy using HAR dataset with ALA.}
    \label{fig:har_sia_ala}
\end{minipage}
\hfill
\begin{minipage}[b]{0.49\columnwidth}
    \centering
    \includestandalone[width=\linewidth]{fig/HAR_ALA/EOD} 
    \caption{Equal opportunity difference (EOD) using HAR dataset with ALA.}
    \label{fig:har_EOD_ala}
\end{minipage}
\end{figure}

\subsection{Differential Privacy (DP) with HAR}\label{appendix:har_dp}

Figure~\ref{fig:accu_har_DP} illustrates the training and testing accuracy for HAR when using DP-SGD, while Figures~\ref{fig:har_SIA_DP} and~\ref{fig:har_loss_DP} show the FI(SIA) and FI(loss).

\begin{figure}[!h]
  \centering
  \begin{subfigure}[b]{0.49\columnwidth}
    \centering
    \includestandalone[width=\textwidth]{fig/HAR-DP_FINP_ALA/train}
    \caption{Training accuracy. }
  \end{subfigure}
  \hfill
  \begin{subfigure}[b]{0.49\columnwidth}
    \centering
    \includestandalone[width=\textwidth]{fig/HAR-DP_FINP_ALA/test}
     \caption{Testing accuracy.}
  \end{subfigure}
  \caption{Global model classification accuracy using HAR with different levels of noise in DP-SGD.}
  \label{fig:accu_har_DP}
\end{figure}


\begin{figure}[!h]
\centering
\begin{minipage}[b]{0.49\columnwidth}
\includestandalone[width=\linewidth] {fig/HAR-DP_FINP_ALA/FI_SIA}
    \caption{\small{Disparity of SIA accuracy (FI(SIA)) among clients using HAR dataset with  different noise in DP-SGD.}}
  \label{fig:har_SIA_DP}
  \end{minipage}
  \hfill
 \begin{minipage}[b]{0.49\columnwidth}
 \includestandalone[width=\linewidth] {fig/HAR-DP_FINP_ALA/FI_loss}
  \caption{\small{Disparity of prediction loss FI(Loss) among clients using HAR and different levels of noise in DP-SGD.}}
  \label{fig:har_loss_DP}
 \end{minipage}
\end{figure}

\begin{figure}[!t]
\centering
\begin{minipage}[b]{0.49\columnwidth}
    \centering
    \includestandalone[width=\linewidth]{fig/HAR-DP_FINP_ALA/SIA_average_acc}
    \caption{Average SIA accuracy using HAR dataset with ALA and different noise in DP-SGD.}
    \label{fig:har_sia_dp}
\end{minipage}
\hfill
\begin{minipage}[b]{0.49\columnwidth}
    \centering
    \includestandalone[width=\linewidth]{fig/HAR-DP_FINP_ALA/EOD} 
    \caption{EOD using HAR dataset with ALA and different noise in DP-SGD.}
    \label{fig:har_EOD_dp}
\end{minipage}
\end{figure}

\subsection{Correlation between the top Hessian eigenvalue ($\lambda_{\text{max}}$) and Hessian trace ($H_{T}$)}
\label{hessian_appendix}

\begin{figure*}[!ht]
\centering
\includegraphics[scale=0.8, width=0.7\linewidth] {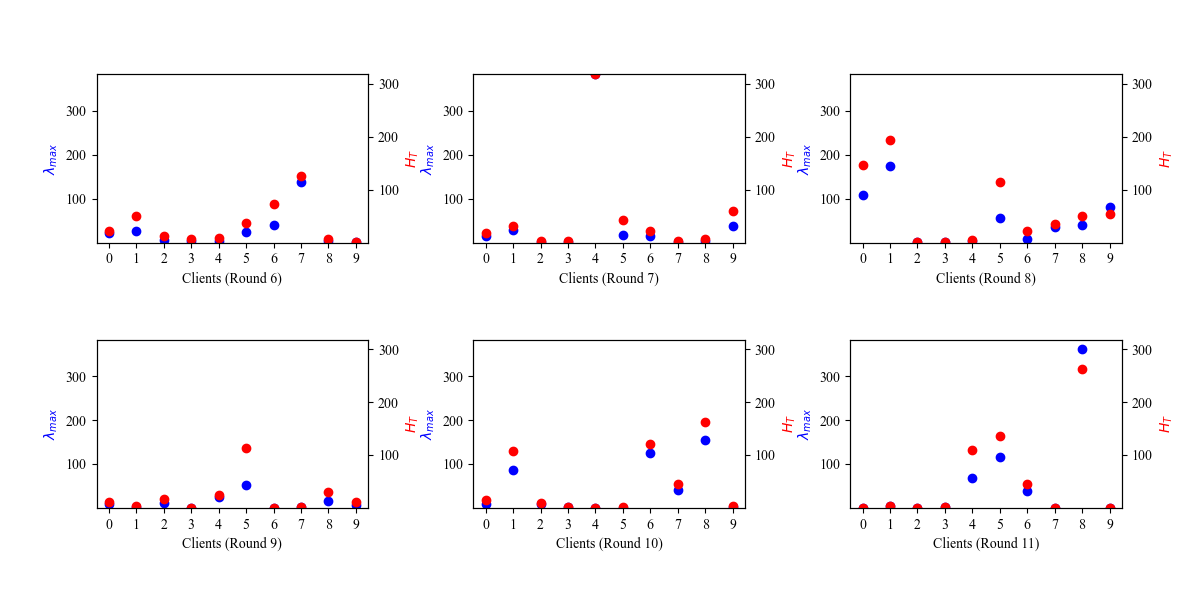}
\caption{Strong correlation between the top Hessian eigenvalue ($\lambda_{\text{max}}$, blue dots) and Hessian trace ($H_{T}$, red dots) across training rounds in HAR.}
\label{fig:Hessian}
\end{figure*}

Figure~\ref{fig:Hessian} shows the strong correlation between the top Hessian eigen value ($\lambda_{\text{max}}$) and the Hessian trace ($H_{T}$).

\newpage
\section{Experiment Results on CIFAR-10}
\label{appendix cifar}
Complete results for ablation experiments on $\beta$, ALA and PCA comparison, and DP-SGD can be found in \autoref{tbl:cifar_ablation}.

\begin{table*}[!t]
\small
\caption{Results of CIFAR dataset using ResNet as the local model. }
\centering

\centering
\begin{tabularx}{0.97\linewidth}{|c|c|c||c|c||c|c|c||c||}
\hline
\multicolumn{1}{|l|}{} & \multicolumn{2}{c||}{Accuracy (\%) } & \multicolumn{2}{c||}{Privacy Metrics (\%))} & \multicolumn{3}{c||}{Fairness Metrics}& Efficiency \\ \hline\hline

& Train & Test & \makecell{Mean  (SIA$_{acc}$) $\downarrow$} & \makecell{Max (SIA$_{acc}$)}$\downarrow$  & \makecell{CoV(SIA$_{acc}$)/FI(SIA$_{acc}$)} & \makecell{CoV(loss)/FI(Loss)} & EOD $\downarrow$ & \makecell{Conv. round} \\ \hline
Baseline & 78.39 & 76.45 & 30.86 & 38.52 & \textbf{0.33/0.90} & 0.67/0.69 & 0.33 &  10\\
FedAlign~\cite{mendieta2022local} & 70.79 & 71.87 & 30.72 & 38.46 & 0.34/0.89 & 0.86/0.58 & 0.33 &  14\\

\makecell{\sysname (with PCA)} & \textbf{80.23} & \textbf{78.47} & \textbf{10.07} & \textbf{10.67} & 0.35/0.89 & \textbf{0.44/0.83} & \textbf{0.12} &  12\\

\hline

\end{tabularx}

\label{tbl:cifar_resnet}
\end{table*}

\subsection{Equal Opportunity Difference (EOD) and SIA accuracy}
Figures~\ref{fig:cifar sia cnn ablation} and~\ref{fig:cifar EOD cnn ablation} illustrate the average SIA accuracy in CIFAR-10 with CNN across different $\beta$, and the EOD results respectively.

\begin{figure}[!h]
\centering
\begin{minipage}[b]{0.49\columnwidth}
    \centering \includestandalone[width=\linewidth]{fig/CIFAR-PETS/line_SIA_Accuracy}
    \caption{Average SIA accuracy of Baseline and different $\beta$ using CIFAR-10 with CNN.}
    \label{fig:cifar sia cnn ablation}
\end{minipage}
\hfill
\begin{minipage}[b]{0.49\columnwidth}
    \centering    \includestandalone[width=\linewidth]{fig/CIFAR-PETS/line_EOD} 
    \caption{Equal opportunity difference (EOD) using CIFAR-10 with CNN.}
    \label{fig:cifar EOD cnn ablation}
\end{minipage}
\end{figure}

\begin{table*}[!t]
\small
\caption{Ablation experiments on $\beta$. Results of CIFAR dataset using CNN as the local model~\citep{hu2023source}. }
\centering
\begin{tabularx}{0.96\linewidth}{|c|c|c||c|c||c|c|c||c||}
\hline
\multicolumn{1}{|l|}{} & \multicolumn{2}{c||}{Accuracy (\%) } & \multicolumn{2}{c||}{Privacy Metrics (\%)} & \multicolumn{3}{c||}{Fairness Metrics}& Efficiency \\ \hline\hline

& Train & Test & \makecell{Mean  (SIA$_{acc}$) $\downarrow$} & \makecell{Max (SIA$_{acc}$)}$\downarrow$  & \makecell{CoV(SIA$_{acc}$)/FI(SIA$_{acc}$)} & \makecell{CoV(loss)/FI(Loss)} & EOD $\downarrow$ & \makecell{Conv.round} \\ \hline

Baseline~\cite{hu2023source} & 75.62 & 62.37 & 40.91 & 46.70 & 0.23/0.95 & 0.46/0.83 & 0.29 &   5\\\hline

\textbf{\sysname} &&&&&&&& \\

($\beta$=0.05, PCA) & 69.31 & 59.67 & 39.51 & 43.40 & 0.21/0.96 & 0.46/0.83 & 0.27 &  5\\

($\beta$=0.1, PCA) & 70.45 & 61.19 & 39.47 & 43.70 & \textbf{0.19/0.96} & 0.44/0.84 & \textbf{0.25} &  7\\

($\beta$=0.3, PCA) & 71.17 & 63.81 & \textbf{31.85} & \textbf{39.90} & 0.31/0.90 & \textbf{0.38/0.87} & 0.31 &   12\\


($\beta$=0.5, PCA) & 10 & 10 & N/A & N/A &  N/A & N/A & N/A
& N/A\\
 ($\beta$=0.05, ALA) & \textbf{76.03} & \textbf{64.26} & 38.62 & 43.90 & 0.23/0.95 & 0.46/0.83 & 0.29 &  6 \\
($\beta$=0.1, ALA) & 74.64 & 63.94 & 37.39 & 42.50 & 0.25/0.94 & 0.44/0.84 & 0.32 &  7 \\
 ($\beta$=0.3, ALA) & 71.00 & 64.09 & 34.09 & 41.00 & 0.34/0.89 & 0.41/0.86 & 0.38 &  11 \\

\hline
\textbf{DP-SGD} ($\epsilon,\delta$) &&&&&&&& \\
(93, $10^{-5}$) & 43.73 & 43.59 & 23.38 & 24.50 &  0.52/ 0.79& 0.290/0.922  & 0.37  & 15 \\
(16, $10^{-5}$) & 42.84 & 42.62 & 22.68 & 24.20 & 0.45/0.83 &  0.298/0.918 & 0.31 & 15 \\
(5, $10^{-5}$) &31.52  & 32.16 & 21.85 & 24.60 & 0.38/0.88 & 0.271/0.931 & 0.29 & 13 \\

\hline

\end{tabularx}
\label{tbl:cifar_ablation}
\end{table*}

\subsection{Differential Privacy (DP) with CIFAR-10} 
\label{appendix:cifar_dp}

Figure~\ref{fig:accu cifar cnn DP} highlights the global model classification accuracy in CIFAR-10 with CNN under different levels of noise in DP-SGD. Figures~\ref{fig:CIFAR_SIA_DP} and~\ref{fig:CIFAR_loss_DP} show the disparity of SIA accuracy (FI(SIA)) and disparity of prediction loss (FI(loss)) among clients using CIFAR-10 with CNN and under different noise in DP-SGD, respectively. Figures~\ref{fig:cifar_sia_dp} and~\ref{fig:cifar_EOD_dp} illustrate the average SIA accuracy and EOD respectively using ALA in CIFAR-10 with CNN and different noise in DP-SGD

\begin{figure}[!h]
  \centering
  \begin{subfigure}[b]{0.49\columnwidth}
    \centering
    \includestandalone[width=\textwidth]{fig/CIFAR-DP/train_accuracy}
    \caption{Training accuracy. }
  \end{subfigure}
  \hfill
  \begin{subfigure}[b]{0.49\columnwidth}
    \centering
    \includestandalone[width=\textwidth]{fig/CIFAR-DP/test_accuracy}
     \caption{Testing accuracy.}
  \end{subfigure}
  \caption{Global model classification accuracy using CIFAR-10 with CNN and different levels of noise in DP-SGD.}
  \label{fig:accu cifar cnn DP}
\end{figure}

\begin{figure}[!h]
\centering
\begin{minipage}[b]{0.49\columnwidth}
\includestandalone[width=\linewidth] {fig/CIFAR-DP/FI_SIA}
    \caption{\small{Disparity of SIA accuracy (FI(SIA)) among clients using CIFAR-10 dataset with CNN and different noise in DP-SGD.}}
  \label{fig:CIFAR_SIA_DP}
  \end{minipage}
  \hfill
 \begin{minipage}[b]{0.49\columnwidth}
 \includestandalone[width=\linewidth] {fig/CIFAR-DP/FI_loss}
  \caption{\small{Disparity of prediction loss FI(Loss) among clients using CIFAR-10 with CNN and different levels of noise in DP-SGD.}}
  \label{fig:CIFAR_loss_DP}
 \end{minipage}
\end{figure}

\begin{figure}[!t]
\centering
\begin{minipage}[b]{0.49\columnwidth}
    \centering
    \includestandalone[width=\linewidth]{fig/CIFAR-DP/SIA_average_acc}
    \caption{Average SIA accuracy in CIFAR-10 dataset using ALA with CNN and different noise in DP-SGD.}
    \label{fig:cifar_sia_dp}
\end{minipage}
\hfill
\begin{minipage}[b]{0.49\columnwidth}
    \centering
    \includestandalone[width=\linewidth]{fig/CIFAR-DP/EOD} 
    \caption{EOD in CIFAR-10 dataset using ALA with CNN and different noise in DP-SGD.}
    \label{fig:cifar_EOD_dp}
\end{minipage}
\end{figure}

\section{Experiment Results on FEMNIST}\label{app:feminist}

\subsection{Additional Fairness and Disparity Metrics}

We utilize fairness notions beyond simple average performance and incorporate metrics that explicitly quantify the distributional disparity among clients. Specifically, we utilize the Mean Absolute Difference (MAD) and Sen's Social Welfare Function.

\textbf{Mean Absolute Difference (MAD)}
The Mean Absolute Difference quantifies the average absolute disparity between all possible pairs of individuals or clients within the system. For a population of size $n$, where $x_i$ represents the metric evaluated for client $i$, the MAD is defined as:

\begin{equation}
    \text{MAD} = \frac{1}{n^2} \sum_{i=1}^{n} \sum_{j=1}^{n} |x_i - x_j|
\end{equation}

Unlike variance, which squares differences and thus heavily weights extreme outliers, MAD provides a linear, highly interpretable measure of overall systemic inequality. Crucially, assuming $x_i \in [0, 1]$, MAD reaches its mathematical upper bound of $0.5$ not under a monopoly (where one client holds all utility), but under conditions of \textit{perfect polarization}---where exactly half of the population receives a score of $1$ and the other half receives $0$. Therefore, MAD is exceptionally well-suited for identifying models that systematically disenfranchise specific clusters or demographics while perfectly serving others and independent of the mean values.

\textbf{Sen's Social Welfare Function} To holistically assess the system, it is necessary to balance overall efficiency with equity. Sen's Social Welfare Function \cite{sen1997economic} achieves this by scaling the global average performance ($\mu$) by an inequality penalty. By fully expanding the Gini coefficient ($G$) within the standard formulation $W = \mu(1 - G)$, the function is explicitly defined as:

\begin{equation}
    W = \mu \left( 1 - G \right) = \mu \left( 1 - \frac{\sum_{i=1}^{n} \sum_{j=1}^{n} |x_i - x_j|}{2n^2 \mu} \right)
\end{equation}

This formula calculates the egalitarian equivalent performance of the system, assuming the target metric (e.g., utility or predictive accuracy) is one where a higher value is optimal. It bridges the gap between raw average performance and distributional fairness. 

\begin{itemize}[leftmargin=*]
    \item \textbf{The Global Mean ($\mu$):} The outer multiplier represents the system's overall average performance. The function inherently rewards pushing this global average as high as possible. \textit{In our context of attack accuracy, where a lower value is optimal, we calculate the $\mu$ based on (1-$Accuracy_{attack}$).}
    \item \textbf{The Expanded Disparity Penalty:} The fractional term inside the parentheses is the expanded Gini coefficient. The double summation in the numerator ($\sum \sum |x_i - x_j|$) exhaustively calculates the absolute disparity between every possible pair of clients in the system. The denominator ($2n^2 \mu$) normalizes this disparity relative to the total number of pairwise comparisons and the size of the mean itself.
\end{itemize}

By subtracting this normalized disparity from $1$ and multiplying the result by the mean, the function imposes a strict structural penalty on inequality. The system cannot achieve a high Welfare score simply by maximizing performance for a privileged subset of clients (which would inflate $\mu$ but also drastically inflate the disparity penalty). Instead, it is mathematically forced to distribute performance as equitably as possible across all clients.

\subsection{Hessian Calculation Overhead}

To demonstrate the practical viability of deploying FinP in real-world federated environments, we evaluate the computational overhead introduced by the Hessian calculations during our FEMNIST experiments. Because our framework relies primarily on the dominant (top) Hessian eigenvalue instead of full Hessian Matrix, and it does not require exact numerical precision to effectively rank client vulnerability, we can aggressively optimize the solver to significantly reduce the computational burden. 

In our experiments, the global neural network architecture has a lightweight memory size of 3.275 MB CNN. To optimize for speed, we configure the eigensolver with highly relaxed convergence constraints: a maximum of 20 iterations for both the eigenvalue solver and the trace estimator (\texttt{hessian\_eig\_max\_iter=20}, \texttt{hessian\_trace\_max\_iter=20}), alongside a tolerance threshold of $5 \times 10^{-3}$ (\texttt{hessian\_tol=5e-3}). Under these lightweight hyperparameters, the dominant Hessian eigenvalue computation requires only 1.82 seconds per client. This minimal overhead confirms that \sysname\ can be seamlessly integrated into standard federated training pipelines without imposing a prohibitive computational bottleneck on edge devices.

\begin{figure}[!t]
  \centering
  \begin{subfigure}[b]{0.49\columnwidth}
    \centering
    \includestandalone[width=\textwidth]{fig/FEMNIST/femin_line_average_train_accuracy}
    \caption{\scriptsize{Global model training accuracy.}}
    \label{fig:accCV}
  \end{subfigure}
  \hfill
  \begin{subfigure}[b]{0.49\columnwidth}
    \centering
    \includestandalone[width=\textwidth]{fig/FEMNIST/femin_line_average_test_accuracy}
    \caption{\scriptsize{Global model testing accuracy.}}
    \label{fig:femnist_accFI}
  \end{subfigure}
  \caption{Global model classification accuracy using FEMNIST dataset. \sysname\ (ALA) maintains competitive accuracy with minimal convergence delay.}
  \label{fig:femnistaccu}
\end{figure}

\begin{figure}[!t]
\centering
\begin{minipage}[b]{0.49\columnwidth}
\includestandalone[width=\linewidth] {fig/FEMNIST/SIA_FI}
  \caption{\small{Disparity of SIA accuracy (FI(SIA)) among clients using FEMNIST dataset with \sysname\ with ALA.}}
  \label{fig:FEMNIST sia fi}
  \end{minipage}
  \hfill
\begin{minipage}[b]{0.49\columnwidth}
  \includestandalone[width=\linewidth] {fig/FEMNIST/LOSS_FI}
  \caption{\small{Disparity of prediction loss FI(loss) among clients using FEMNIST dataset with \sysname\ with ALA.}}
  \label{fig:FEMNIST loss fi}
  \end{minipage}
\end{figure}

\begin{figure}[!t]
\centering
\vspace{-2mm}
    \centering
    \includestandalone[scale=0.5]{fig/FEMNIST/feminist_line_SIA_Accuracy}
    \caption{Average SIA accuracy in FEMNIST with \sysname\ with ALA.}
    \label{fig:femnist-sia}
\end{figure}

\begin{figure}[!t]
\centering
\vspace{-2mm}
    \centering
    \includestandalone[scale=0.5]{fig/FEMNIST/loss_mad}
    \caption{Loss MAD (mean absolute difference of loss) across clients in FEMNIST with \sysname\ with ALA.}
    \label{fig:femnist-loss-mad}
\end{figure}

\begin{figure}[!t]
\centering
\vspace{-2mm}
    \centering
    \includestandalone[scale=0.5]{fig/FEMNIST/welfare}
    \caption{Sen's welfare across clients in FEMNIST with \sysname\ with ALA.}
    \label{fig:femnist-welfare}
\end{figure}

\section{Detailed Experimental Setup and Results for Per-Round White-Box MIA}
\label{appendix:mia}

To evaluate the vulnerability of individual client updates, we simulate a per-round gradient-matching Membership Inference Attack (MIA). The adversary is defined as the “honest-but-curious" central server, which has access to the global model parameters $G_{t-1}$ and intercepts the raw, unaggregated local weight updates $\Delta w_i$ from each client $i$ at round $t$.

\subsection{Data Partitioning and Threat Model Setup}
We partition the overall dataset to ensure strict separation between the adversary's auxiliary knowledge and the final evaluation data. The federated cohort consists of $N$ participating clients.

\begin{itemize}
    \item \textbf{Dummy Client (Client $0$):} The adversary injects one sybil client into the federated process. This client participates in aggregation at every round, serving as the ``Member'' baseline.
    \item \textbf{Victim Clients (Clients $1$ to $N-1$):} The genuine participants who is targeted in MIA.
    \item \textbf{Shadow Non-Member Pool ($\mathcal{D}_{shadow}$):} A pool of data (e.g., an isolated writer) held by the adversary, exclusively used to generate ``Non-Member'' features for classifier training.

    \item \textbf{Evaluation Sets ($\mathcal{D}_{test}$):} For each victim client $i$, we construct a static evaluation dataset of $100$ records: $50$ randomly sampled records from client $i$'s local data (Label $1$) and $50$ records from unseen writer data (Label $0$).
\end{itemize}

\subsection{Dynamic Feature Generation (Round $t$)}
Because full-participation FL continuously memorizes data over time, static MIA thresholds fail. Instead, the adversary calibrates a dynamically trained Random Forest classifier prior to the aggregation step at each round $t$. We generate a balanced training set of $60$ feature vectors ($30$ positive, $30$ negative) using a bootstrapping approach on the Dummy Client.

For each of the $30$ iterations, we generate a paired sample:
\begin{enumerate}
    \item \textbf{Target Sampling:} The adversary samples a member record $x_m \sim \text{Client } 0$ and a non-member record $x_{nm} \sim \mathcal{D}_{shadow}$.
    \item \textbf{Gradient Extraction:} Using standard Cross-Entropy Loss, the adversary computes the individual gradient footprints $g_m$ and $g_{nm}$ with respect to the current global model $G_{t-1}$.
    \item \textbf{Mock Local Update Simulation:} The adversary samples a random $80\%$ subset of Client 0's local dataset, ensuring the inclusion of $x_m$. A temporary clone of $G_{t-1}$ is trained on this subset using the standard local hyperparameters to yield a mock update $\Delta w_{mock}$.
    \item \textbf{Feature Representation:} To map the high-dimensional parameter space to a robust feature space, the adversary calculates the layer-wise cosine similarity. The positive feature vector is $\cos(g_m, \Delta w_{mock})$, and the negative feature vector is $\cos(g_{nm}, \Delta w_{mock})$.
\end{enumerate}

\subsection{Classifier Training and Inference}
The $60$ layer-wise similarity vectors are used to train an Random Forest Classifier. This classifier serves as the decision boundary for round $t$.

To execute the attack, the adversary intercepts the real uploaded update $\Delta w_i$ from each victim client $i \in \{1, \dots, N-1\}$. For each record $x_{eval}$ in the victim's $100$-record Evaluation Set $\mathcal{D}_{test}$, the adversary computes the gradient $g_{eval}$ on $G_{t-1}$ and extracts the layer-wise cosine similarity against the intercepted $\Delta w_i$. The similarity vector is passed to the Random Forest to predict membership. We compute standard classification metrics (Accuracy, Precision, Recall, and ROC-AUC) against the ground-truth labels.

\subsection{MIA Results}

our evaluations demonstrate that the FinP framework inherently provides an effective defense against Membership Inference Attacks (MIAs). Specifically, we observe that MIA accuracy drops by an average of $6.1\%$ under the FinP setup compared to the baseline in \autoref{fig:femnist-mia} and \autoref{tbl:FEMNISTresults-MIA}. Although the MIA accuracy disparity across clients are minor (FI = 0.99), \sysname\ consistently decreases the disparity of loss (MAD) from 0.638 to 0.574, and decreases the EOD by 0.22 among clients.  This empirical reduction highlights that FinP actively mitigates privacy leakage by fostering a more generalized model training process, thereby reducing the localized overfitting and instance-level memorization that gradient-matching MIAs rely upon to succeed.

\begin{table*}[!t]
\small
\setlength{\tabcolsep}{3pt}
\caption{Results of FEMNIST dataset with MIA attack.}
\centering
\resizebox{\linewidth}{!}{%
\begin{tabular}{|l|c|c||c|c||c|c|c|c|c||c||}
\hline
 & \multicolumn{2}{c||}{Accuracy (\%)} & \multicolumn{2}{c||}{Privacy Metrics (\%)} & \multicolumn{5}{c||}{Fairness Metrics} & Efficiency \\ \hline\hline
 & Train & Test & \makecell{Mean\\(MIA$_{acc}$)$\downarrow$} & \makecell{Max\\(MIA$_{acc}$)$\downarrow$} & \makecell{CoV(MIA$_{acc}$)\\/FI(MIA$_{acc}$)} & \makecell{CoV(loss)\\/FI(Loss)} & \makecell{EOD \\(MIA)$\downarrow$} & \makecell{MAD\\(loss)$\downarrow$} & \makecell{Welfare \\(MIA)$\uparrow$} & \makecell{Conv.\\round} \\ \hline\hline
Baseline~\cite{hu2023source}     & 72.72 & 65.04 & 66.63 & 73.11 & 0.088/0.99 & 0.281/0.916 & 0.40 & 0.638 & 0.302 & 10 \\ \hline
\sysname\ ($\beta=1$)            & 58.65 & 60.38 & \textbf{60.53} & \textbf{70.56} & 0.099/0.990 & \textbf{0.215/0.949} & 0.21 & 0.606 & 0.304 & 12 \\
\sysname\ ($\beta=0.75$)         & 65.23 & 62.34 & 66.57 & 75.22 & \textbf{0.089/0.991} & 0.229/0.942 & \textbf{0.18} & \textbf{0.574} & 0.302 & 12 \\
\sysname\ ($\beta=0.5$)          & \textbf{68.03} & \textbf{63.76} & 86.31 & 74.78 & 0.093/0.990 & 0.257/0.928 & 0.20 & 0.608 & 0.283 & 14 \\ 
\hline
DP-SGD ($\epsilon,\delta$)     &       &       &       &       &             &             &      &       &       & \\
\quad (151, $10^{-5}$)                      & 49.57 & 50.10 & 50.58 & 58.22 & 0.038/0.996 & 0.288/0.920 & 0.06 & 2.062 & 0.483 & 14 \\
\quad (92, $10^{-5}$)                   & 27.01 & 29.01 & 50.11 & 52.44 & 0.038/0.998 & 0.257/0.938 & 0.06 & 2.005 & 0.488 & - \\
\quad (31, $10^{-5}$)                     &  2.95 &  2.50 & 13.62 & 16.56 & 0.041/0.997 & 0.436/0.838 & 0.07 & 10.097 & 0.493 & - \\
\hline
\end{tabular}%
}
\label{tbl:FEMNISTresults-MIA}
\end{table*}

\begin{figure}[!t]
\centering
\vspace{-2mm}
    \centering
    \includestandalone[scale=0.5]{fig/FEMNIST/mia}
    \caption{MIA attack accuracy on FEMNIST with \sysname\ with ALA.}
    \label{fig:femnist-mia}
\end{figure}

\begin{figure}[!t]
\centering
\vspace{-2mm}
    \centering
    \includestandalone[scale=0.5]{fig/FEMNIST/femin_line_MIA_Precision}
    \caption{MIA precision on FEMNIST with \sysname\ with ALA.}
    \label{fig:femnist-mia-precision}
\end{figure}

\begin{figure}[!t]
\centering
\vspace{-2mm}
    \centering
    \includestandalone[scale=0.5]{fig/FEMNIST/femin_line_MIA_Recall}
    \caption{MIA recall on FEMNIST with \sysname\ with ALA.}
    \label{fig:femnist-mia-recall}
\end{figure}

\begin{figure}[!t]
\centering
\vspace{-2mm}
    \centering
    \includestandalone[scale=0.5]{fig/FEMNIST/femin_line_MIA_ROC_AUC}
    \caption{MIA ROC-AUC on FEMNIST with \sysname\ with ALA.}
    \label{fig:femnist-mia-rocauc}
\end{figure}

\end{document}